%% file: main.tex
\documentclass[10pt]{article}
\input{packages}

\input{preamble}

\title{World-Ego Modeling for Embodied~Video Generation in Long-Horizon Navigation-Manipulation Tasks}

\input{author}

\begin{document}

\maketitle

\input{Sections/0_abs}
\input{Sections/1_intro}
\input{Sections/2_relate}
\input{Sections/3_method}
\input{Sections/4_exp}
\input{Sections/5_con}

\bibliographystyle{unsrtnat}
\bibliography{main}

\newpage
\appendix
\input{Sections/X_app}

\end{document}

%% file: packages.tex
\usepackage{hmi-preprint}
\usepackage{amsmath}
\usepackage{url}
\usepackage{booktabs}
\usepackage{nicefrac}
\usepackage{microtype}
\usepackage{marvosym}
\usepackage{xcolor}         
\definecolor{cvprblue}{rgb}{0.21,0.49,0.74}
\usepackage[table]{xcolor}
\usepackage{enumitem}
\usepackage{amsmath}
\usepackage{multirow}
\hypersetup{
    colorlinks=true, 
    linkcolor=cvprblue, 
    citecolor=cvprblue, 
    urlcolor=cvprblue,  
}

%% file: preamble.tex
\preprintlogos{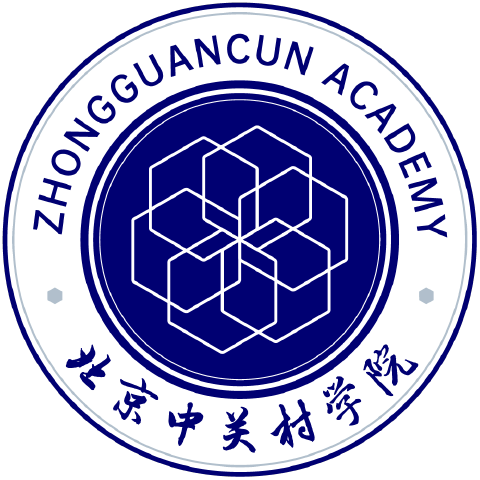}{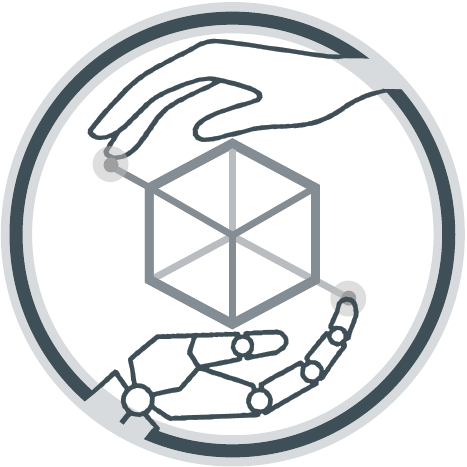}

\preprintdate{May 2026}

\runningtitle{World-Ego Modeling for Long-Horizon Evolution in Hybrid Embodied Tasks}

\preprintfooter{\textsuperscript{\Letter} Corresponding author.}

\preprintlinkset
  {https://zgca-hmi-lab.github.io/WEM}
  {https://github.com/ZGCA-HMI-Lab/WEM}
  {https://huggingface.co/Zoorao/WEM}
  {https://huggingface.co/datasets/Zoorao/HTEWorld}

\preprintteaser{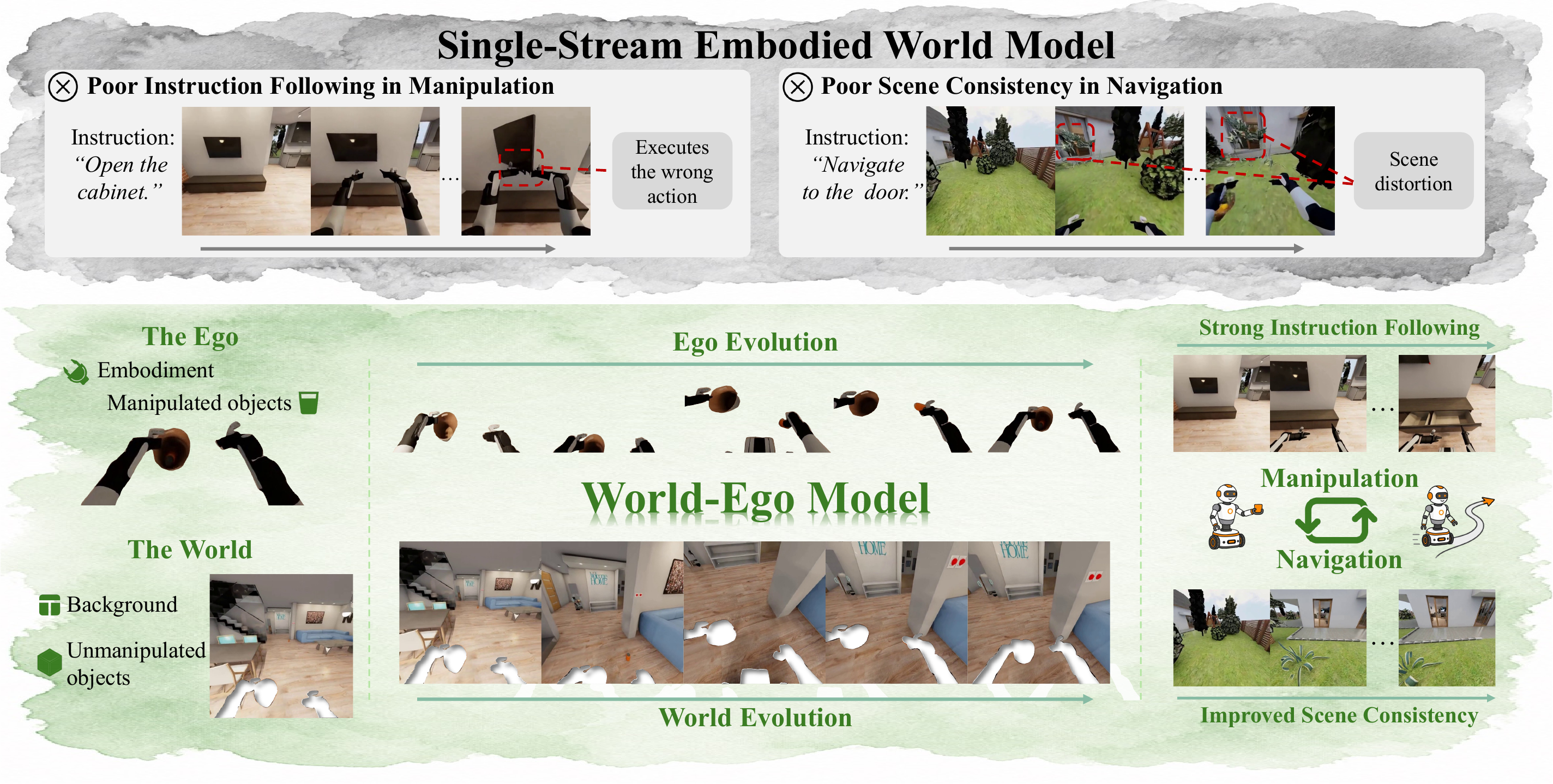}{
Motivation of WEM.
Embodied world models usually generate video
in a single stream, leading to difficulties in accurate instruction following and long-horizon scene consistency. 
We define the world as the background and currently unmanipulated objects, and the ego as the robot and currently manipulated objects. 
Further, we separately model the evolution of the world and ego, thereby improving video generation performance in long-horizon navigation-manipulation tasks.
}

%% file: author.tex
\author{
Zuyao Lin\textsuperscript{1,2,3} \hspace{0.15in}
Jianhui Zhang\textsuperscript{3,4} \hspace{0.15in}
Peidong Jia\textsuperscript{5} \hspace{0.15in}
Xiaoguang Zhao\textsuperscript{1} \hspace{0.15in} \\ 
Shanghang Zhang\textsuperscript{5} \hspace{0.15in}
Jingdong Wang \textsuperscript{3} \hspace{0.15in}
Xingyu Chen \textsuperscript{3 \Letter}
}

\affil{
\textsuperscript{1}Institute of Automation, Chinese Academy of Sciences \hspace{0.15in} \\
\textsuperscript{2}University of Chinese Academy of Sciences \hspace{0.15in} \\
\textsuperscript{3}Zhongguancun Academy \hspace{0.15in}
\textsuperscript{4}Shanghai Jiaotong University \hspace{0.15in}
\textsuperscript{5}Peking University
}

%% file: Sections/0_abs.tex
\begin{abstract}
Embodied video world models typically capture both scene evolution and the robot's behavior, which we refer to as the \emph{world} and the \emph{ego}, respectively.
The world and the ego exhibit different underlying dynamics: world prediction relies primarily on visual history and emphasizes scene stability, whereas ego prediction relies more strongly on the current instruction and emphasizes accurate instruction following.
Modeling both components within a single generation stream can entangle these different dependencies, making it difficult to specialize the prediction of either component.
Consequently, it becomes difficult to simultaneously maintain scene consistency and accurate instruction following, particularly in long-horizon navigation-manipulation tasks.
In this paper, we propose to decompose an embodied video into the world and the ego and disentangle their generation processes.
Specifically, we define the world as the background and currently unmanipulated objects, and the ego as the robot and currently manipulated objects.
Based on this definition, we develop the \emph{World-Ego Model} (WEM), which combines a vision-language state predictor using role-conditioned attention (RCA) and asymmetric query budgets with 
a semantic-routed mixture-of-experts (SR-MoE) diffusion generator.
To enable rigorous evaluation, we further construct HTEWorld, a dataset and benchmark for long-horizon embodied video generation with hybrid navigation-manipulation tasks, providing 125K training video clips comprising over 4.5M frames with fine-grained instructions, together with 300 multi-turn evaluation trajectories covering over 2K instructions.
Extensive experiments show that WEM achieves state-of-the-art performance on HTEWorld while remaining competitive on existing manipulation-oriented evaluations.
\end{abstract}

%% file: Sections/1_intro.tex
\section{Introduction}
\label{sec:introduction}

World models are essential to embodied AI, as they learn physical dynamics to predict future consequences, generate synthetic data, and serve as policy simulators~\citep{bib:world_models, bib:dreamerv3, bib:dreamgen, bib:cosmos_policy}.
Recent video-based world models have demonstrated strong capabilities in generating realistic future rollouts and are playing an increasingly important role in embodied intelligence~\citep{bib:cosmos, bib:robodreamer, bib:pan}. 

An embodied video can be viewed as comprising two components: the \emph{world}, which captures the evolution of the environment, and the \emph{ego}, which captures the behavior of the embodiment and its interactions with objects.
Existing embodied world models generally do not explicitly distinguish the world from the ego. 
Instead, they use a single generation stream to jointly model scene evolution, robot behavior, and object interactions. 
They have shown strong performance in short-horizon and manipulation-oriented prediction~\citep{bib:robodreamer,bib:tesseract,bib:worldarena}. 
However, when applied to more complex tasks, such as long-horizon hybrid navigation-manipulation involving interleaved navigation and manipulation stages, they struggle to maintain scene consistency \citep{bib:longscape} or follow instructions~\citep{bib:rbench} accurately, as shown in \autoref{fig:teaser}.

We attribute this challenge partly to mixing the world and the ego within a single generation stream. 
The world and the ego exhibit different 
underlying dynamics, i.e., world evolution relies primarily on historical context, whereas ego evolution relies more strongly on the current instruction~\citep{bib:worldagen}.
Modeling both within a single generation stream entangles these different dependencies and hinders specialization for either component. This motivates us to explicitly separate the conditioning information and generation parameters associated with the world and the ego.

As illustrated in \autoref{fig:teaser}, we explicitly decompose an embodied video into world and ego components, defining the world as the background and currently unmanipulated objects and the ego as the robot and currently manipulated objects.
Based on this decomposition, we introduce the \textbf{World-Ego Model} (\textbf{WEM}), which separates world and ego information for prediction and introduces specialized experts for generation.
WEM has prediction and generation stages.
In the prediction stage, a VLM-based state predictor~\citep{bib:qwen3-vl} extracts world and ego states from the multimodal context using role-conditioned attention (RCA) and asymmetric query budgets.
In the generation stage, we introduce a Semantic-Routed Mixture-of-Experts (SR-MoE) generator with a shared Preceding Expert and specialized World/Ego Experts.
The Preceding Expert first processes all video tokens, after which a Semantic Head predicts their world-ego assignments. The assigned world and ego tokens are then routed to their corresponding specialized experts, whose outputs are merged to produce the next video chunk.
Combined with chunk-wise autoregressive diffusion~\citep{bib:magi}, WEM supports long-horizon multi-turn generation for hybrid navigation-manipulation tasks.

Our setting requires evaluating both navigation-oriented scene imagination and manipulation-oriented physical simulation under continuous multi-turn embodied instructions, but none of the recent benchmarks~\citep{bib:vbench++,bib:worldmodelbench, bib:rbench, bib:worldarena} meet our requirements.
We therefore construct \textbf{Hybrid-Task Embodied World Benchmark} (\textbf{HTEWorld}) on top of BEHAVIOR-1K~\citep{bib:behavior}.
HTEWorld provides 125K training video clips containing over 4.5M frames with fine-grained instructions, together with 300 evaluation trajectories containing over 2K instructions and interleaved navigation-manipulation stages.
WEM achieves SOTA performance on HTEWorld and remains competitive under existing manipulation-oriented tasks~\citep{bib:worldarena,bib:droid}.
Our contributions are summarized as follows:

\begin{itemize}[itemsep=5pt, parsep=2pt, topsep=0pt]
    \item We show that an embodied video can be decomposed into world (i.e., background and currently unmanipulated objects) and ego (i.e., the robot and currently manipulated objects) components, providing an effective formulation for embodied video world models.

    \item We develop WEM to instantiate world-ego decomposition with a world-ego state predictor and an SR-MoE video generator. We demonstrate its effectiveness for long-horizon, multi-turn generation in hybrid navigation-manipulation tasks.

    \item We establish HTEWorld, which provides 125K training clips and 300 evaluation trajectories, as a dataset and benchmark for long-horizon embodied video generation in hybrid navigation-manipulation tasks. Experiments show that WEM achieves SOTA performance on HTEWorld while remaining competitive on existing manipulation-oriented evaluations.
\end{itemize}

%% file: Sections/2_relate.tex
\vspace{-7pt}
\section{Related Work}
\vspace{-7pt}

\paragraph{Video World Models.}
World models predict future states from historical observations, actions, or instructions, serving as internal simulators for planning, data generation, and policy learning~\citep{bib:world_models,bib:dreamer,bib:dreamerv3,bib:daydreamer}. Early methods learn compact latent dynamics for latent imagination. With diffusion models~\citep{bib:ddpm,bib:image_ldm} and video generation~\citep{bib:sora, bib:wan, bib:cogvideox, bib:genie2, bib:open_sora, bib:vid2world, bib:matrix2, bib:hunyuanvideo, bib:motionctrl, bib:flovd, bib:rays_as_pixels, bib:motion_attribution}, future prediction has moved from low-dimensional state transitions to pixel-level visual rollouts. Cosmos-Predict~\citep{bib:cosmos_predict} builds a post-trainable video world foundation model for Physical AI, supporting future observation prediction and synthetic data generation. Recent works further advance interactive video world models: Genie~\citep{bib:genie} learns a latent action model from unlabeled videos; The Matrix~\citep{bib:the_matrix}, Yume~\citep{bib:yume}, and LIVE~\citep{bib:live} target real-time controllable interaction, open-ended scene exploration, and long-horizon consistency modeling; PAN~\citep{bib:pan} proposes a generative latent prediction framework for general, interactable, and long-horizon world simulation conditioned on history and language actions.

\vspace{-9pt}
\paragraph{Embodied Video World Models.}
Embodied world models predict both scene evolution and how agent behaviors change the physical world. Existing methods predict future robot observations via video generation~\citep{bib:unisim, bib:robodreamer, bib:tesseract, bib:roboscape} or action-conditioned rollouts~\citep{bib:bridgedata_v2}. Such predictions support downstream uses, including data generation~\citep{bib:astra_mobile_robots}, policy learning~\citep{bib:motus, bib:anypoint_trajectory}, evaluation~\citep{bib:worldarena}, control~\citep{bib:cosmos_policy}, and video-level planning~\citep{bib:large_video_planner}. WoW~\citep{bib:wow} learns a generative world model from real robot trajectories and uses inverse dynamics to translate imagined outcomes into executable actions. Ctrl-World~\citep{bib:ctrl_world} builds a controllable multi-view world model for robot manipulation, using pose-conditioned memory retrieval and frame-level action conditioning for long-horizon policy imagination, evaluation, and improvement. 
However, most embodied video world models use a single generation stream to jointly model scene evolution, robot behavior, and object interactions, despite their different conditioning dependencies. Related works have explored structured modeling of visual dynamics from different perspectives. 
Iso-Dream~\citep{bib:iso_dream} separates controllable and noncontrollable dynamics for model-based reinforcement learning, 
while GEM~\citep{bib:gem} introduces explicit controls for ego motion and object dynamics in egocentric video generation.
In contrast, we decompose embodied videos into world and ego components according to their current interaction roles and separately model their evolution for long-horizon hybrid navigation-manipulation generation.

%% file: Sections/3_method.tex
\vspace{-7pt}
\section{Method}
\label{sec:method}
\vspace{-7pt}

An embodied video world model predicts a future video $\mathbf{V}$ from the available context $\mathcal{H}$ and the current control signal $a$ according to $\mathbf{V}\sim p_\theta(\mathbf{V}\mid\mathcal{H},a)$, where $\mathcal{H}$ may contain the initial observation, previously generated videos, and past control signals, while $a$ denotes the current control signal, such as a text instruction, camera trajectory, or robot action.

\input{Figures/fig4_model}

\subsection{World-Ego Decomposition}
\label{subsec:world_ego_modeling}

Existing embodied video world models typically use a single generation stream.
However, different parts of an embodied video follow different conditioning dependencies. The background and objects not involved in the current interaction are primarily constrained by the observed history and should preserve scene structure over time. In contrast, the robot and currently manipulated objects are driven by the current control signal and must realize the intended behavior. Asking a single predictive mapping to model these two dependencies makes it difficult to specialize for either one.

We therefore decompose an embodied video into a \textbf{world} component and an \textbf{ego} component. Specifically, the world consists of the background and currently unmanipulated objects, while the ego consists of the robot and currently manipulated objects; 
conceptually,
$\mathbf{V}=\mathcal{C}(\mathbf{V}^{w},\mathbf{V}^{e})$, where $\mathcal{C}$
denotes their composition.
WEM realizes this decomposition at two complementary
levels.
The \textbf{world-ego state predictor} separates the conditioning information required by the two roles, producing world and ego states with different contextual dependencies~(\autoref{subsec:world_ego_predictor}); the \textbf{SR-MoE video generator} retains shared video processing while introducing specialized generation parameters for world and ego regions~(\autoref{subsec:world_ego_generator}). As illustrated in \autoref{fig:model}, these two stages are applied autoregressively to generate a sequence of video chunks for long-horizon generation.

\subsection{World-Ego State Predictor}
\label{subsec:world_ego_predictor}

The world and ego require different information from the generation history: the world benefits from long-term context to maintain persistent scene content across rounds, whereas the ego depends on the most recent visual observation and the current instruction to perform the interaction at the current round.
To accommodate these different information requirements, we extract two role-specific feature groups from the same multimodal context, one for the world and one for the ego.

For the $k$-th generation round, we construct the multimodal context from the initial frame $\mathbf{O}_0$, previous instruction-video pairs $(a_1,\mathbf{V}_1),\ldots,(a_{k-1},\mathbf{V}_{k-1})$, and the current instruction $a_k$. These inputs are arranged chronologically and encoded into a vision-language token sequence $\mathbf{X}_k\in\mathbb{R}^{L_k\times d_v}$, where $L_k$ is the sequence length and $d_v$ is the VLM hidden dimension.
To extract separate world and ego states from this shared sequence, we append two groups of learnable queries, $\mathbf{Q}^w\in\mathbb{R}^{N_w\times d_v}$ and $\mathbf{Q}^e\in\mathbb{R}^{N_e\times d_v}$, where $N_w$ and $N_e$ denote the numbers of world and ego queries, respectively. 
A pretrained VLM $\Phi_\phi$~\citep{bib:qwen3-vl} processes $\mathbf{X}_k$ together with these queries and produces
\begin{equation}
\left(
\mathbf{S}^w_k,
\mathbf{S}^e_k
\right)
=
\Phi_\phi
\left(
\mathbf{X}_k;
\mathbf{Q}^w,\mathbf{Q}^e
\right).
\end{equation}
Here, $\mathbf{S}^w_k\in\mathbb{R}^{N_w\times d_v}$ and $\mathbf{S}^e_k\in\mathbb{R}^{N_e\times d_v}$ are the VLM outputs at the world and ego query positions, which we refer to as the world and ego states, respectively.

\vspace{-9pt}
\paragraph{Role-Conditioned Attention.}
Using different query tokens alone does not guarantee different information: if both query groups attend to the same context, they may capture largely overlapping information. We therefore explicitly control the context accessible to each query group through Role-Conditioned Attention (RCA).
Specifically, both query groups extract information from the generation history, but with different attention scopes.
The world queries attend to the complete historical vision-language context and to one another, while being blocked from the current instruction and the ego queries. This biases the world state toward stable scene information, rather than letting it directly depend on the current instruction. In contrast, the ego queries attend to the most recent video chunk, the current instruction, and one another, while being blocked from earlier history and the world queries. This focuses the ego state on the interaction to be executed and reduces interference from stale information. 
As a result, RCA encourages $\mathbf{S}^w_k$ and $\mathbf{S}^e_k$ to capture complementary information for world and ego generation while retaining a shared VLM backbone.
As illustrated in \autoref{fig:model}, RCA is implemented by reconstructing the VLM attention mask.

\paragraph{Asymmetric Query Budgets.}
The two roles also differ in the amount of context they must summarize. Since the world state aggregates a substantially longer and more diverse history, we allocate more queries to the world than to the ego. 
In practice, this is implemented by setting different numbers of learnable queries, $N_w$ and $N_e$, which provides a flexible way to adjust the capacity allocated to the two states.
This provides the world state with greater representation capacity while keeping the ego state compact and focused on the current interaction.

\subsection{SR-MoE Video Generator}
\label{subsec:world_ego_generator}

Separating the conditioning states alone does not fully resolve world-ego entanglement if both regions are still modeled with the same generation parameters. However, completely separating the two generation streams would sacrifice the shared spatial and temporal structure required for coherent video synthesis. 
We therefore combine shared processing for common video structure with specialized processing for the world and ego.
We restructure a pretrained video DiT~\citep{bib:dit} into a Semantic-Routed Mixture-of-Experts (SR-MoE). The first 24 of the 30 DiT blocks form a shared Preceding Expert, while the final six are duplicated to initialize independently optimized World and Ego Experts. The shared blocks preserve global spatiotemporal interaction across the entire video, whereas the specialized blocks process world and ego tokens with separate parameters.

\paragraph{Preceding Expert.}
Let $\mathbf{Z}_k\in\mathbb{R}^{T\times N_s\times d_g}$ denote the noisy video latent tokens for generation round $k$, where $T$ is the number of frames, $N_s$ is the number of spatial tokens per frame, and $d_g$ is the token dimension. For simplicity, we omit the diffusion timestep from the notation. Before semantic routing, all tokens in $\mathbf{Z}_k$ are jointly processed by the Preceding Expert. This shared processing is important because the world and ego are not independent visual layers: their appearance, geometry, occlusion, and motion must remain coherent within the same video. Each block additionally includes a state cross-attention branch conditioned on the concatenated world and ego states $[\mathbf{S}^w_k;\mathbf{S}^e_k]$, allowing the Preceding Expert to incorporate information from both states during shared processing. We denote the output of its $\ell$-th block as $\mathbf{Z}^{(\ell)}_k$, and its final output as $\mathbf{Z}^P_k\in\mathbb{R}^{T\times N_s\times d_g}$.

\paragraph{Neighbor-Expanded Assignment.}
After shared processing, the tokens need to be assigned to the World or Ego Expert for specialized processing. Since the robot and manipulated objects move across frames, we dynamically predict the world-ego assignment using selected intermediate outputs of the Preceding Expert and the ego state. Specifically, the Semantic Head is implemented as a dense prediction transformer~\citep{bib:dpt} that takes the intermediate outputs $\{\mathbf{Z}^{(\ell)}_k\}_{\ell\in\mathcal{J}}$ from a selected set of blocks $\mathcal{J}$ together with the ego state $\mathbf{S}^e_k$ and predicts \begin{equation} \mathbf{M}_k = \sigma\left( G_{\mathrm{sem}} \left( \{\mathbf{Z}^{(\ell)}_k\}_{\ell\in\mathcal{J}}, \mathbf{S}^e_k \right) \right) \in[0,1]^{T\times N_s}, \end{equation} where $G_{\mathrm{sem}}$ denotes the Semantic Head and $\sigma$ denotes the sigmoid function. The resulting $\mathbf{M}_k$ is a frame-wise ego probability mask aligned with the video, with each element $M_{k,t,i}$ indicating the probability that spatial token $i$ in latent frame $t$ belongs to the ego region. 
Given a threshold $\tau$, we define the world and ego index sets as $\mathcal{I}^w_k=\{(t,i)\mid M_{k,t,i}<\tau\}$ and $\mathcal{I}^e_k=\{(t,i)\mid M_{k,t,i}\geq\tau\}$.

A naive hard assignment scheme would send each token exclusively to one expert. This creates two related problems near the boundary separating world and ego tokens. Spatially, each expert loses access to neighboring tokens assigned to the other role, even though their visual content must remain locally coherent. Temporally, as the robot or manipulated objects move, the role assigned to a given spatial location can change across adjacent frames, causing that location to switch abruptly between independently optimized experts and potentially introducing temporal discontinuities such as flickering near the boundary. We therefore introduce Neighbor-Expanded Assignment. 
Rather than treating the predicted assignment as a strict separation between the two experts, we create a narrow overlapping region around the boundary. 
For each role $r\in\{w,e\}$, we expand its routing index set as $\widetilde{\mathcal{I}}^r_k=\mathcal{I}^r_k\cup\mathcal{N}(\mathcal{I}^r_k)$, where $\mathcal{N}(\mathcal{I}^r_k)$ denotes the indices of spatial neighbors within the same frame for positions indexed by $\mathcal{I}^r_k$.
The expansion allows tokens close to the boundary to be seen by both experts, while tokens sufficiently far from the boundary remain role-specific. 
It therefore provides cross-boundary spatial context and reduces abrupt changes in token routing across adjacent frames by allowing tokens near the boundary to be processed by both experts.

\paragraph{Routing, Role Experts, and Unrouting.}
The expanded token groups are gathered from the Preceding Expert output and routed to their corresponding experts: \begin{equation} \mathbf{Z}^r_k = \operatorname{Gather}\left( \mathbf{Z}^P_k, \widetilde{\mathcal{I}}^r_k \right), \qquad \widehat{\mathbf{Z}}^r_k = E^r_{\theta_r}\left( \mathbf{Z}^r_k, a_k, \mathbf{S}^r_k \right), \qquad r\in\{w,e\}. \end{equation} Here, $\mathbf{Z}^r_k,\widehat{\mathbf{Z}}^r_k\in\mathbb{R}^{|\widetilde{\mathcal{I}}^r_k|\times d_g}$ denote the routed tokens before and after expert processing, respectively, and $E^w_{\theta_w}$ and $E^e_{\theta_e}$ denote the World and Ego Experts. Each expert contains a state cross-attention branch conditioned on its corresponding state $\mathbf{S}^r_k$. After expert processing, the expert outputs are restored to their original spatiotemporal positions. 
For positions covered by only one expanded routing index set, the corresponding expert output is used directly; for positions shared by both index sets, the two outputs are averaged.
We denote this operation as \begin{equation} \widehat{\mathbf{Z}}_k = \operatorname{Restore} \left( \widehat{\mathbf{Z}}^w_k,\widetilde{\mathcal{I}}^w_k; \widehat{\mathbf{Z}}^e_k,\widetilde{\mathcal{I}}^e_k \right) \in\mathbb{R}^{T\times N_s\times d_g}. \end{equation} This design preserves shared processing for global video coherence while allowing the World and Ego Experts to specialize according to the predicted world-ego assignment.

\paragraph{Training.}
WEM is trained end-to-end with a flow-matching objective~\citep{bib:flow_matching} for video generation and an auxiliary mask-prediction objective for the Semantic Head. The latter combines class-balanced binary cross-entropy and Dice losses~\citep{bib:dice}. The overall objective is $\mathcal{L}=\mathcal{L}_{\mathrm{flow}}+\lambda\left(\mathcal{L}_{\mathrm{BCE}}+\mathcal{L}_{\mathrm{Dice}}\right)$, where $\lambda$ weights the mask-prediction loss.

%% file: Figures/fig4_model.tex
\begin{figure}[t]
\begin{center}
\includegraphics[width=\linewidth]{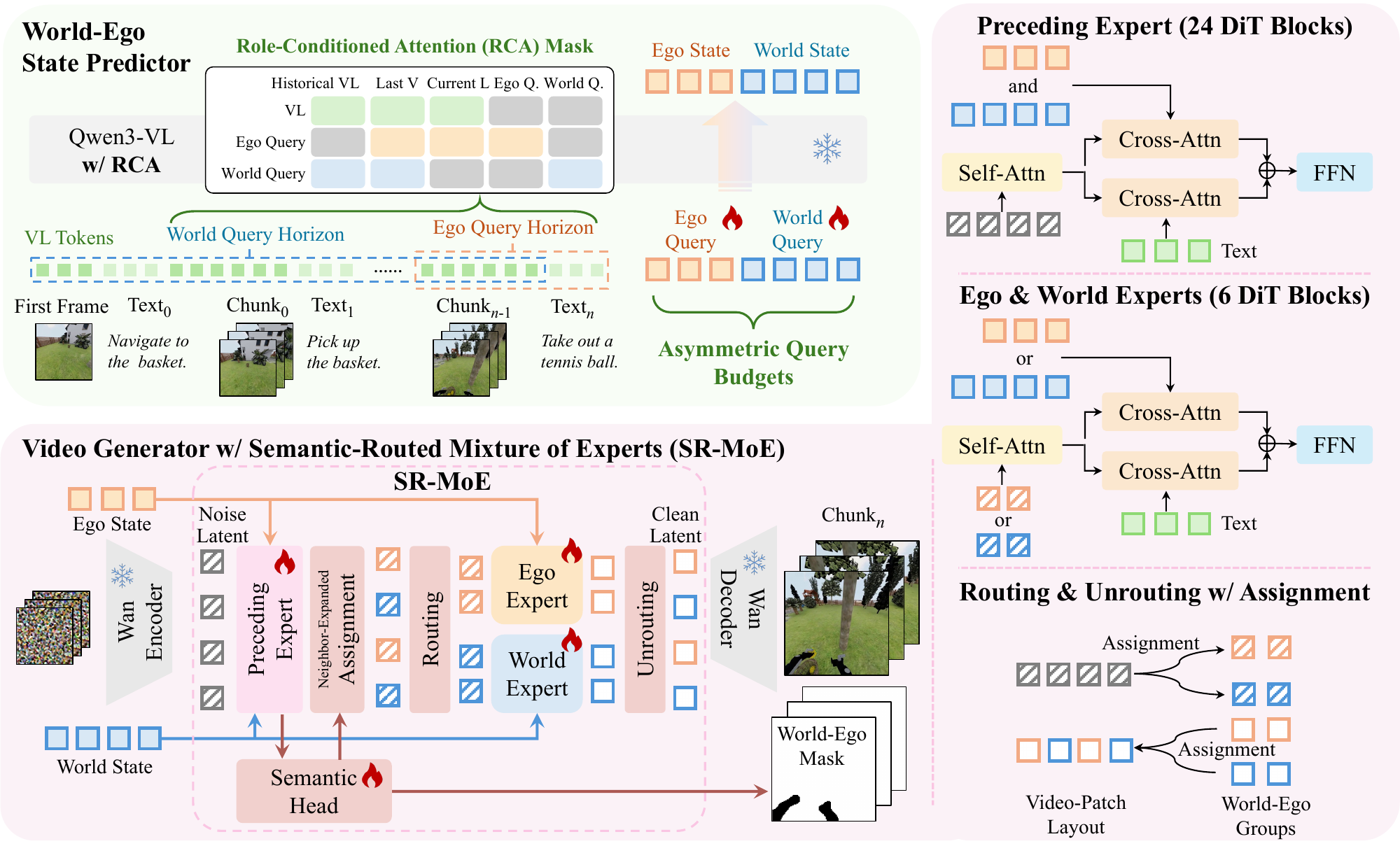}
\caption{
Overview of WEM architecture.
A VLM-based state predictor uses asymmetric world and ego query groups with RCA to extract world and ego states from the vision-language inputs.
The SR-MoE generator consists of a shared Preceding Expert and specialized World and Ego Experts.
The Preceding Expert processes the noisy video latent tokens, and a Semantic Head predicts their world-ego assignment.
The assigned tokens are routed to the corresponding experts, whose outputs are restored to their original positions.
}
\label{fig:model}
\end{center}
\end{figure}

%% file: Sections/4_exp.tex
\section{HTEWorld}
\label{subsec:hteworld}

\input{Figures/dataset}

Existing embodied video generation benchmarks mainly focus on short video clips or individual instructions~\citep{bib:vbench,bib:vbench++,bib:vbench2,bib:worldmodelbench,bib:rbench,bib:worldarena}. They do not adequately evaluate long-horizon generation involving interleaved navigation and manipulation.
We therefore construct HTEWorld from the teleoperated demonstrations released for the 2025 BEHAVIOR Challenge in the BEHAVIOR-1K environment~\citep{bib:behavior}. The dataset provides household-task demonstrations collected with the Galaxea R1 Pro robot and multimodal observations.
We select ten representative tasks and use the head-camera RGB observations from the demonstrations to construct HTEWorld. For each task, five episodes are held out for evaluation, while the remaining episodes are used for training.

The videos are segmented into 37-frame clips at $16$ FPS and a resolution of $480\times480$. 
HTEWorld contains approximately 127K video clips, including 125K training clips and 2K evaluation clips. The evaluation set contains 300 multi-turn trajectories with interleaved navigation and manipulation stages.
\autoref{fig:hteworld_dataset} summarizes the statistics of HTEWorld. Details of the data construction and annotation pipeline are provided in Appendix~\autoref{app:dataset}.
We adopt the video quality evaluation protocol introduced by WorldArena~\citep{bib:worldarena}, which comprises 16 metrics across six dimensions: visual quality, motion quality, content consistency, physics adherence, 3D accuracy, and controllability. We report the aggregated EWMScore as the primary metric.
We additionally introduce six HTEWorld-specific metrics for multi-turn consistency and navigation-manipulation transitions, detailed in \autoref{app:metrics}.

\section{Experiments}
\label{sec:experiments}

\subsection{Experimental Setup}
\label{subsec:setup}

We build the state predictor on Qwen3-VL-2B-Instruct~\citep{bib:qwen3-vl}. We empirically set the query budgets to $N_w=192$ for the world and $N_e=64$ for the ego.
The SR-MoE video generator is initialized from either Wan2.1-T2V-1.3B or Wan2.2-TI2V-5B~\citep{bib:wan}, resulting in WEM (Wan2.1) and WEM (Wan2.2), respectively. We freeze the Qwen3-VL backbone and jointly optimize all remaining WEM parameters. We set the routing threshold $\tau$ to 0.5 and anneal the mask-loss weight $\lambda$ from 0.3 to 0.06 over training, emphasizing semantic-assignment learning early and video generation later. WEM is trained for four epochs with a learning rate of $1\times10^{-5}$. Both variants are evaluated on HTEWorld, while all other experiments use WEM (Wan2.2).

We compare WEM with Cosmos-Predict2.5-2B and Cosmos-Predict2.5-14B~\citep{bib:cosmos_predict}, WoW-7B~\citep{bib:wow}, and two PAN-style baselines~\citep{bib:pan}. PAN (Wan2.1) and PAN (Wan2.2) use the same Qwen3-VL and Wan initializations as their corresponding WEM variants, but predict a unified state and retain a single generation stream. 
Unless otherwise specified, all methods are trained on the same data and evaluated under the same protocol for each benchmark.
Complete implementation, training, and evaluation details are provided in \autoref{app:eval_protocol}.

\subsection{Main Results}
\label{subsec:hteworld_results}

\input{Tables/tab_hteworld_worldarena}

\input{Figures/qualitative_exp}

\paragraph{Evaluation on HTEWorld.}
As shown in \autoref{tab:hteworld_worldarena}, WEM performs strongly on HTEWorld for long-horizon multi-turn generation in hybrid navigation-manipulation tasks. WEM (Wan2.2) and WEM (Wan2.1) achieve the highest and second-highest EWMScores of 61.48 and 60.11, respectively.
More importantly, the gains concentrate on the two capabilities that motivate world-ego decomposition: scene consistency and instruction following. Compared with PAN (Wan2.2), WEM (Wan2.2) improves background consistency (BC) by 7.68 points and subject consistency (SC) by 7.28 points, showing substantially more stable scene evolution. It also improves semantic alignment (Sem) by 3.07 points, instruction following (Inst) by 1.60 points, and interaction quality (Inter) by 4.47 points, indicating better adherence to the instructed behavior.
The qualitative results in \autoref{fig:qualitative_rollout} show the same pattern: while the compared models gradually suffer from scene distortion or fail to follow later instructions, WEM better preserves the scene and completes the full instruction sequence.
These improvements are consistent across different video generation backbones: WEM outperforms its PAN counterpart by 3.53 points with Wan2.1 and by 3.08 points with Wan2.2, demonstrating the effectiveness of world-ego decomposition across different video-generation backbones. Results under the HTEWorld-specific metrics are reported in \autoref{app:metrics_results}.

\paragraph{Evaluation on WorldArena.}
We further evaluate WEM on WorldArena~\citep{bib:worldarena}, which focuses on short-horizon tabletop manipulation. As shown in \autoref{tab:worldarena_results}, WEM achieves an EWMScore of 58.10, the highest score among the compared text-conditioned methods. Using text instructions alone, it also performs on par with methods using stronger action-based control signals. 
These results show that world-ego decomposition remains effective for manipulation-only tasks, where separating the persistent scene from robot-object interactions is still beneficial.

\input{Tables/tab_worldarena_droid_alternatives}

\paragraph{Evaluation on DROID.}
We evaluate WEM on DROID~\citep{bib:droid}, a large-scale real-world dataset covering diverse scenes and robot manipulation tasks. As shown in \autoref{tab:droid_results}, WEM (Wan2.2) achieves an EWMScore of 53.70, outperforming PAN (Wan2.2) by 2.87 points and the Wan2.2 baseline by 7.26 points. These results extend our findings from simulated environments to real-world robot videos, showing that world-ego decomposition remains effective under the greater visual diversity and less controlled conditions of real-world data.

\subsection{Ablation Studies}
\label{subsec:ablations}

\autoref{tab:wem_ablation} shows our ablation studies on state prediction, video generation, and training strategy.

\input{Tables/tab_wem_ablations}

\paragraph{Effect of world-ego state predictor.}
Comparing rows 3 and 1 of \autoref{tab:wem_ablation}, introducing a unified state predictor improves EWMScore from 53.49 to 58.40 under the same single-stream generator without mask supervision.
This substantial improvement shows the benefit of explicitly predicting a state to condition future generation. 
Comparing rows 1 and 5, replacing the unified state with separate world and ego states further improves the score to 58.69, with the generator and training scheme unchanged.
This controlled comparison shows that separating the conditioning information required by the world and ego roles provides an additional benefit beyond state prediction itself.

We further examine the state predictor's Role-Conditioned Attention and asymmetric query budget using full WEM with the Wan2.2 video-generation backbone (61.48 EWMScore). Removing Role-Conditioned Attention and allowing both query groups to access the complete multimodal context reduces the score to 60.64. Replacing the 192 world and 64 ego queries with 128 queries for each role reduces the score to 60.50. These results support allocating different capacities and attention scopes to the two states, consistent with their different contextual dependencies described in \autoref{subsec:world_ego_predictor}.

\paragraph{Effect of SR-MoE.}
Comparing rows 3 and 4 of \autoref{tab:wem_ablation}, replacing the single-stream generator with SR-MoE improves EWMScore from 53.49 to 57.71 without the state predictor. Comparing rows 5 and 6, the same replacement improves the score from 58.69 to 61.48 with world-ego states. Since SR-MoE also introduces mask supervision, we examine this factor separately by comparing rows 1 and 2: adding the same auxiliary supervision to PAN improves its score only from 58.40 to 58.61. This smaller gain suggests that auxiliary mask supervision alone does not explain the improvements associated with SR-MoE, supporting the benefit of role-specific generation.

\input{Tables/tab_sr_moe_ablation}

We further examine the key design choices of SR-MoE in \autoref{tab:alternative_designs}. Under semantic assignment, the full SR-MoE achieves an EWMScore of 61.48 (row 6). Removing token separation reduces the score to 61.09 (row 5), while removing expert specialization decreases it to 58.85 (row 4). These results show that both token separation and expert specialization contribute to the effectiveness of SR-MoE.
To validate the semantic assignment used in SR-MoE, we compare it with a motion-based alternative that identifies ego regions from motion induced by the robot and manipulated objects, as illustrated in \autoref{fig:views}. Comparing rows 1 and 4, 2 and 5, and 3 and 6, semantic assignment improves EWMScore by 0.08, 1.73, and 4.44 points, respectively.
This advantage can be attributed to semantic masks being directly tied to visual appearance and therefore easier to predict reliably from video latents, as shown in \autoref{fig:role_expert_visualization}, where the predicted assignments remain accurate across the generated sequence.
The SR-MoE variants are illustrated in \autoref{fig:framework}, with implementation details provided in \autoref{app:variants}.

We additionally examine Neighbor-Expanded Assignment within SR-MoE using the Wan2.2 video-generation backbone. Removing it reduces EWMScore from 61.48 to 59.57 and produces visible discontinuities in the generated videos. This result supports providing overlapping context near the boundary between world- and ego-assigned regions, consistent with the motivation in \autoref{subsec:world_ego_generator}.

%% file: Figures/dataset.tex
\begin{figure*}[t]
\centering
\includegraphics[width=\textwidth]{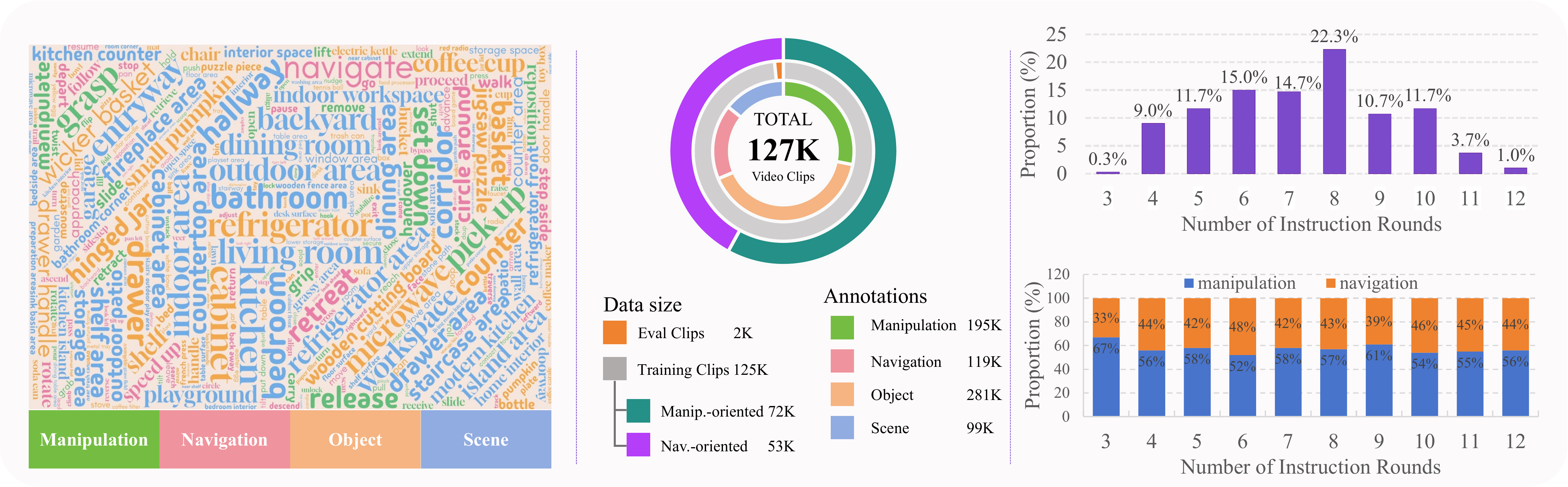}
\caption{
Statistics of HTEWorld.
HTEWorld contains large-scale training clips and multi-turn evaluation trajectories for hybrid navigation-manipulation generation.
Left: word-frequency visualization of manipulation, navigation, object, and scene annotations.
Middle: dataset composition, including the training/evaluation split, manipulation/navigation training clips, and counts of actions, objects, and scenes mentioned in the instructions.
Right: evaluation-trajectory statistics, showing the distribution of instruction rounds and manipulation/navigation proportions across trajectory lengths.
}
\label{fig:hteworld_dataset}
\end{figure*}

%% file: Tables/tab_hteworld_worldarena.tex
\begin{table*}[t]
\centering
\scriptsize
\setlength{\tabcolsep}{2.6pt}
\caption{\footnotesize\raggedright
\textbf{Comparison on HTEWorld} using the WorldArena metrics and EWMScore.
Metric abbreviations follow WorldArena.
Higher is better.
}
\label{tab:hteworld_worldarena}
\resizebox{\textwidth}{!}{%
\begin{tabular}{l|c|ccc|ccc|ccc|ccc|ccc|c}
\toprule
Model
& EWMScore
& \multicolumn{3}{c|}{Visual}
& \multicolumn{3}{c|}{Motion}
& \multicolumn{3}{c|}{Consistency}
& \multicolumn{3}{c|}{3D}
& \multicolumn{3}{c|}{Control}
& Physics \\
&
& IQ & AQ & JEPA
& DD & Flow & MS
& BC & SC & Photo
& DA & TA & Persp
& Sem & Act & Inst
& Inter \\
\midrule
WoW-7B~\citep{bib:wow}
& 53.44
& 64.72 & 49.74 & 1.30
& 22.76 & 25.49 & 67.74
& 66.86 & 63.06 & 38.08
& 82.41 & 28.16 & 95.14
& 86.75 & 3.47 & 78.42
& 80.90 \\

Cosmos-Predict~2.5-2B~\citep{bib:cosmos_predict}
& 54.83
& 64.40 & 50.21 & 1.26
& 24.62 & 27.23 & 69.33
& 71.54 & 66.88 & \textbf{42.53}
& 82.11 & 28.78 & 95.28
& 87.05 & 3.52 & 79.60
& 83.02 \\

Cosmos-Predict~2.5-14B~\citep{bib:cosmos_predict}
& 55.41
& 62.14 & 50.02 & 1.38
& 29.37 & 32.34 & 71.63
& 73.85 & 68.65 & 35.48
& 82.60 & 28.81 & 94.40
& 87.46 & 3.55 & 80.20
& 84.70 \\

PAN (Wan2.1)~\citep{bib:pan}
& 56.58
& 64.87 & 49.01 & 1.20
& 39.65 & 44.00 & 77.01
& 82.99 & 75.95 & 29.87
& 79.64 & 26.26 & 95.23
& 86.22 & 3.12 & 69.14
& 81.12 \\

PAN (Wan2.2)~\citep{bib:pan}
& 58.40
& 65.48 & 49.00 & 1.70
& 38.14 & 47.43 & 79.47
& 80.24 & 74.79 & 33.15
& 82.39 & 28.75 & 95.08
& 87.67 & 4.42 & 80.40
& 86.33 \\
\midrule

WEM (Wan2.1)
& 60.11
& 65.02 & 49.01 & 1.92
& 38.95 & 47.40 & 82.61
& \textbf{88.19} & 81.95 & 34.22
& 83.69 & 31.91 & 95.33
& 88.49 & 4.02 & \textbf{87.72}
& 81.32 \\

\rowcolor[HTML]{D9E1F2}
WEM (Wan2.2)
& \textbf{61.48}
& \textbf{66.82} & \textbf{50.30} & \textbf{2.49}
& \textbf{41.52} & \textbf{49.21} & \textbf{82.70}
& 87.92 & \textbf{82.07} & 35.95
& \textbf{84.55} & \textbf{34.51} & \textbf{97.60}
& \textbf{90.74} & \textbf{4.50} & 82.00
& \textbf{90.80} \\
\bottomrule
\end{tabular}%
}
\vspace{-12pt}
\end{table*}

%% file: Figures/qualitative_exp.tex
\begin{figure*}[t]
    \centering
    \includegraphics[width=\textwidth]{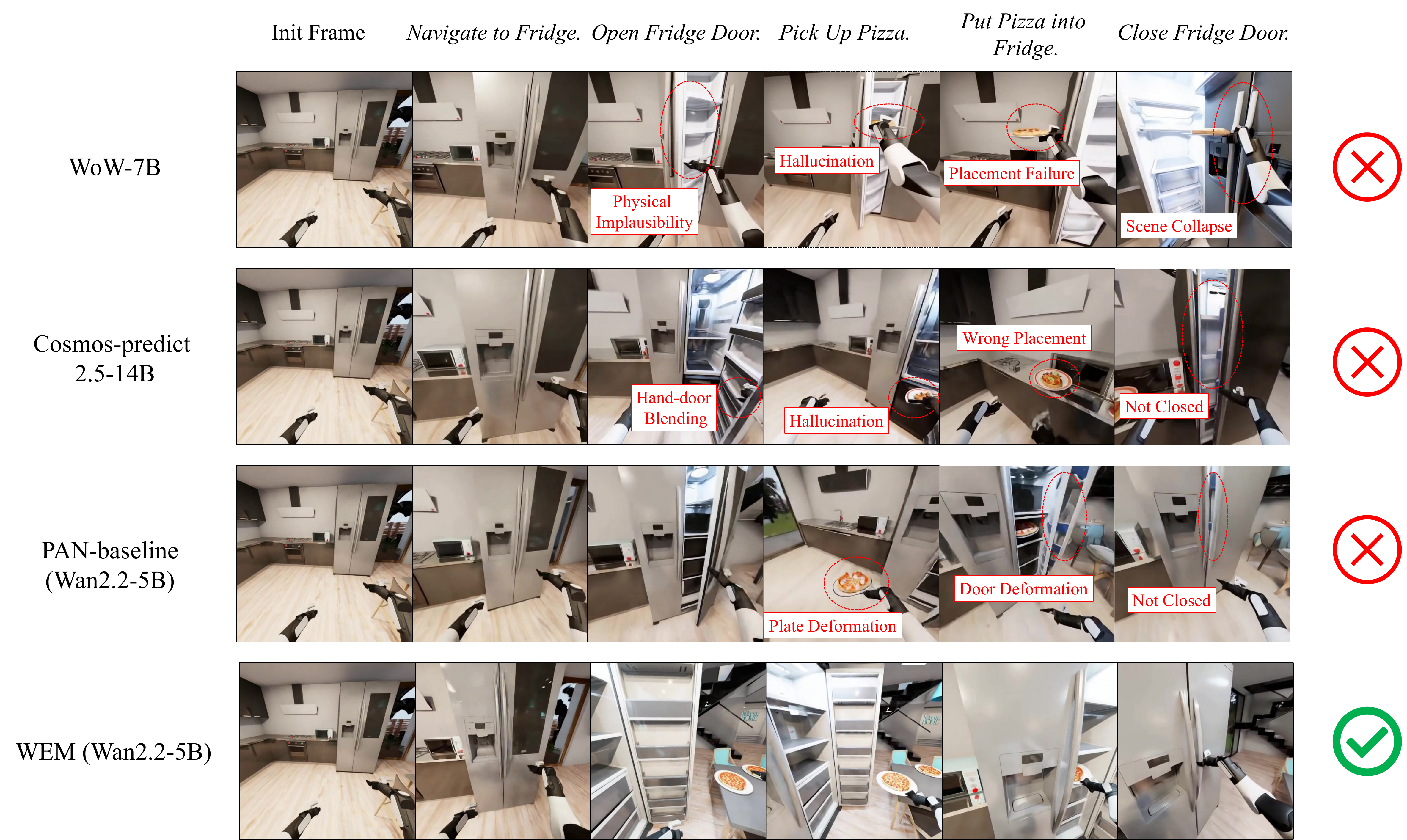}
    \caption{
Qualitative comparison on HTEWorld.
Given the same initial observation and five-step instruction sequence, each model autoregressively generates a long-horizon multi-turn navigation-manipulation rollout.
Red circles indicate scene inconsistency or instruction-following failures. 
WEM 
can better maintain scene consistency and instruction following throughout the rollout.
}
    \label{fig:qualitative_rollout}
\end{figure*}

%% file: Tables/tab_worldarena_droid_alternatives.tex
\begin{table*}[t]
\centering

\begin{minipage}[t]{0.58\textwidth}
\vspace{0pt}
\centering
\scriptsize
\setlength{\tabcolsep}{4pt}
\caption[Comparison on WorldArena.]{\footnotesize\raggedright
\textbf{Comparison on WorldArena.}
The condition denotes the control signal provided in addition to the visual context.
}
\label{tab:worldarena_results}
\resizebox{0.85\linewidth}{!}{%
\begin{tabular}{llc}
\toprule
Model & Condition & EWMScore $\uparrow$ \\
\midrule
Wan2.2~\citep{bib:wan}
& Text & 54.54 \\
WoW~\citep{bib:wow}
& Text & 54.88 \\
CogVideoX~\citep{bib:cogvideox}
& Text & 57.90 \\
\rowcolor[HTML]{D9E1F2}
WEM (Wan2.2)
& Text & 58.10 \\
\midrule
Cosmos-Predict~2.5~\citep{bib:cosmos_predict}
& Action & 55.91 \\
IRASim~\citep{bib:irasim}
& Action & 58.12 \\
Ctrl-World~\citep{bib:ctrl_world}
& Action & \textbf{59.70} \\
\bottomrule
\end{tabular}
}
\end{minipage}
\hfill
\begin{minipage}[t]{0.38\textwidth}
\vspace{0pt}
\centering
\scriptsize
\setlength{\tabcolsep}{4pt}
\caption[Comparison on DROID.]{\footnotesize\raggedright
\textbf{Comparison on DROID.}
PAN and WEM use the Wan2.2 backbone.
}
\label{tab:droid_results}
\resizebox{0.78\linewidth}{!}{%
\begin{tabular}{lc}
\toprule
Model & EWMScore $\uparrow$ \\
\midrule
Wan2.2~\citep{bib:wan}
& 46.44 \\
PAN~\citep{bib:pan}
& 50.83 \\
\rowcolor[HTML]{D9E1F2}
WEM
& \textbf{53.70} \\
\bottomrule
\end{tabular}
}
\end{minipage}
\end{table*}

%% file: Tables/tab_wem_ablations.tex
\begin{table}[t]
\centering
\scriptsize
\setlength{\tabcolsep}{3.2pt}
\caption{\footnotesize\raggedright
\textbf{Ablation of state prediction and generation designs in WEM}, using PAN as the baseline and mask supervision as a control.
Results are reported on HTEWorld.
All variants use Wan2.2~\citep{bib:wan} as the video-generation backbone.
}
\label{tab:wem_ablation}
\resizebox{\linewidth}{!}{%
\begin{tabular}{l|ccc|cc|c|c}
\toprule
\multirow{2}{*}{Model}
& \multicolumn{3}{c|}{State Predictor}
& \multicolumn{2}{c|}{Video Generator}
& Training
& \multirow{2}{*}{EWMScore $\uparrow$} \\
\cmidrule(lr){2-4}\cmidrule(lr){5-6}\cmidrule(lr){7-7}
& None & Unified & World-Ego
& Single Stream & SR-MoE & Mask Supervision
& \\
\midrule
PAN~\citep{bib:pan}
& & $\checkmark$ &
& $\checkmark$ & &
& 58.40 \\
PAN + Mask Supervision
& & $\checkmark$ &
& $\checkmark$ & & $\checkmark$
& 58.61 \\
PAN w/o State Predictor
& $\checkmark$ & &
& $\checkmark$ & &
& 53.49 \\
WEM w/o State Predictor
& $\checkmark$ & &
& & $\checkmark$ & $\checkmark$
& 57.71 \\
PAN + World-Ego States
& & & $\checkmark$
& $\checkmark$ & &
& 58.69 \\
\midrule
\rowcolor[HTML]{D9E1F2}
WEM
& & & $\checkmark$
& & $\checkmark$ & $\checkmark$
& \textbf{61.48} \\
\bottomrule
\end{tabular}%
}
\end{table}

%% file: Tables/tab_sr_moe_ablation.tex
\begin{table}[t]
\centering
\scriptsize
\setlength{\tabcolsep}{5pt}
\caption[SR-MoE design ablations on HTEWorld.]{\footnotesize\raggedright
\textbf{Ablation of SR-MoE design choices on HTEWorld.}
Token Separation indicates whether world and ego tokens are separated according to the predicted assignment before subsequent expert processing.
Expert Specialization indicates whether separate World and Ego Experts are used.
Assignment Proxy specifies whether the world-ego assignment is derived from semantic regions or motion induced by the robot and manipulated objects.
The shaded row denotes the configuration adopted by WEM.
All models use Wan2.2.
}
\label{tab:alternative_designs}
\resizebox{0.80\linewidth}{!}{%
\begin{tabular}{cccc}
\toprule
Token Separation
& Expert Specialization
& Assignment Proxy
& EWMScore $\uparrow$ \\
\midrule
$\checkmark$
&
& Motion
& 58.77 \\

&
$\checkmark$
& Motion
& 59.36 \\

$\checkmark$
& $\checkmark$
& Motion
& 57.04 \\
\midrule

$\checkmark$
&
& Semantic
& 58.85 \\

&
$\checkmark$
& Semantic
& 61.09 \\

\rowcolor[HTML]{D9E1F2}
$\checkmark$
& $\checkmark$
& Semantic
& \textbf{61.48} \\
\bottomrule
\end{tabular}%
}
\end{table}

%% file: Sections/5_con.tex
\section{Conclusion}
\label{sec:conclusion}

This work studies embodied video generation by decomposing a video into world and ego components. We define the world as the background and currently unmanipulated objects, and the ego as the robot and currently manipulated objects. Based on this formulation, we develop WEM, which separates their conditioning information with a VLM-based state predictor and separates their generation parameters with an SR-MoE generator. We construct HTEWorld for long-horizon multi-turn generation with interleaved navigation and manipulation. WEM achieves state-of-the-art performance on HTEWorld while remaining competitive on existing manipulation-oriented evaluations. 

\paragraph{Limitations.}
DROID allows us to evaluate WEM in a real-world tabletop manipulation setting, but the lack of real-world datasets with interleaved navigation and manipulation limits the evaluation of long-horizon hybrid generation in real environments. Moreover, although WEM improves long-horizon generation, errors may still accumulate and earlier context may be lost as generation extends over more instruction rounds or longer durations. A detailed discussion is provided in \autoref{app:limitations}.

%% file: Sections/X_app.tex
\clearpage
\appendix

\section{Data Construction and Annotation}
\label{app:dataset}

\subsection{HTEWorld Data Construction}
\label{app:data_construction}

HTEWorld is constructed from the teleoperated demonstrations released for the 2025 BEHAVIOR Challenge in the BEHAVIOR-1K environment~\citep{bib:behavior}, targeting embodied video generation with interleaved navigation and manipulation. We select ten household tasks: turning on radio; picking up trash; putting away Halloween decorations; cleaning up plates and food; can meat; setting mousetraps; hiding Easter eggs; picking up toys; rearranging kitchen furniture; and setting up a coffee station in the kitchen. These tasks cover diverse navigation and manipulation behaviors, including approaching target locations, opening containers, grasping and carrying objects, and placing objects at designated locations. A representative hybrid sequence is
\emph{navigate to a trash bin $\rightarrow$ pick up the bin $\rightarrow$ navigate to the trash $\rightarrow$ place the trash in the bin}.

The trajectories are recorded from the egocentric head camera of the Galaxea R1 Pro robot. For each task, five episodes are held out for evaluation and excluded from training. We divide the remaining trajectories into fixed-length video clips while preserving the parent action and the temporal position of each clip within that action. Since a single action may span multiple clips, the language annotation describes the behavior visible in each clip rather than directly reusing its parent action label.

Each clip contains 37 frames at 16 FPS and a resolution of 480 $\times$ 480, and incomplete final segments are discarded. The resulting training set contains 125K clips with over 4.5M frames. From the held-out episodes, we select 300 representative multi-turn trajectories with over 2K instructions for evaluation, preserving the ordered navigation and manipulation stages for continuous generation. Table~\ref{tab:app_data_summary} summarizes the resulting data.

\begin{table}[t]
\centering
\small
\caption{Summary of HTEWorld.}
\label{tab:app_data_summary}
\begin{tabular}{ll}
\toprule
Item & Description \\
\midrule
Source environment & BEHAVIOR-1K~\citep{bib:behavior} \\
Embodiment & Wheeled bimanual Galaxea R1 Pro \\
Task coverage & Ten household tasks \\
Training videos & 125K clips; over 4.5M frames \\
Clip format & 37 frames; $480\times480$; 16 FPS \\
Evaluation set & 300 trajectories; over 2K instructions \\
Annotations & Language instructions, semantic masks, optical flow \\
\bottomrule
\end{tabular}
\end{table}

\subsection{Language Instruction Annotation}
\label{app:language_annotation}

We use Gemini 2.5 Flash~\citep{bib:gemini25} to generate one action-centric language instruction for each video clip. The annotation describes the robot behavior and object interaction visible in the current clip rather than directly copying the corresponding task or action label.

The annotation prompt provides the robot embodiment and egocentric viewpoint, the ordered action context of the episode, the current action, and the temporal position of the clip within that action. The Galaxea R1 is described as a wheeled bimanual robot observed from its head camera, with its left and right arms identified by their locations in the image. The action context helps identify objects involved in successive interactions, while the temporal position helps distinguish different stages of a long action, such as approaching an object, manipulating it, and retracting the gripper.

The output is constrained to a single sentence of at most 30 words. It describes observable physical behavior, uses egocentric directions when appropriate, and specifies the acting arm when visible. The model is instructed not to copy the original action label verbatim, infer unobserved future actions, or refer to the recording or annotation process.

Before prompt construction, corrupted action labels are sanitized by removing trailing character sequences introduced by truncated labels. Failed annotation requests are retried, and completed annotations are saved incrementally. The same language annotations are used across all compared world-ego definitions and generator variants.

\subsection{Semantic World-Ego Mask Annotation}
\label{app:semantic_annotation}

The Semantic Head is trained using binary world-ego masks. Following the semantic definition in the main text, the robot and currently manipulated objects are assigned to the ego, while the background and currently unmanipulated objects are assigned to the world. These binary masks provide supervision for the ego probability mask predicted by the Semantic Head.

\paragraph{Instance masks and interaction-dependent roles.}
For HTEWorld, instance segmentation is obtained directly from the BEHAVIOR-1K simulator. Robot pixels are always assigned to the ego. Object instances are assigned according to their role in the current interaction: an object belongs to the world before manipulation, becomes part of the ego while being manipulated or carried, and returns to the world after the interaction. Therefore, instance segmentation determines which pixels belong to each entity, while interaction information determines its current world-ego role.

\paragraph{Interaction-consistency filtering.}
We determine whether a candidate object is currently manipulated using complementary interaction cues, including robot-object overlap, proximity to the end-effector, consistency of motion between robot and object regions, mask confidence, connected-component size, and temporal consistency. Temporal tracking suppresses short-lived or identity-switching assignments and helps preserve object identity through brief occlusions. A candidate object is assigned to the ego only when the available cues consistently support an interaction.

The robot masks and the masks of currently manipulated objects are combined into the ego target, while all remaining pixels are assigned to the world. These annotations are used only for training. During inference, WEM predicts the world-ego assignment internally and does not require external masks.

\subsection{Optical Flow Annotation}
\label{app:flow_annotation}

Optical-flow supervision is used for the motion-based alternative described in Appendix~\ref{app:motion_variant}. This definition requires the optical flow caused by the robot and manipulated objects rather than the total apparent motion in the image.

We estimate dense optical flow $\mathbf{F}$ between consecutive frames using RAFT~\citep{bib:raft}. Camera-induced motion is approximated by fitting a homography $\mathbf{H}$ to matched feature correspondences using RANSAC and converting the estimated transformation into a camera-induced flow field $\mathbf{F}_{\mathrm{cam}}$. We then compute the residual flow as
\begin{equation}
\mathbf{F}_{\mathrm{obj}}
=
\mathbf{F}
-
\mathbf{F}_{\mathrm{cam}},
\label{eq:app_residual_flow}
\end{equation}
which serves as a proxy for the optical flow caused by robot and manipulated-object motion.

The residual flow is approximate because a single homography cannot fully explain viewpoint-induced motion in scenes with varying depth, and additional errors can arise from occlusion and optical-flow estimation. The resulting residual-flow targets are used to supervise the Flow Head in the motion-based variants.

\subsection{Annotation on WorldArena and DROID}
\label{app:external_annotation}

WorldArena~\citep{bib:worldarena} and DROID~\citep{bib:droid} do not provide simulator-exported instance segmentation. We therefore use SAM~3~\citep{bib:sam3} to segment and track the robot and candidate objects in the videos. The resulting masks are processed using the same interaction-consistency procedure described in Appendix~\ref{app:semantic_annotation} to identify currently manipulated objects and construct binary world-ego targets.

Language instructions are generated using the procedure in Appendix~\ref{app:language_annotation}, with the embodiment description and available task context adapted to each dataset. Neither SAM~3 nor Gemini is required during WEM inference.

For DROID, we construct a trajectory-disjoint subset with 5,591 training trajectories (approximately 40K clips) and 294 validation trajectories, corresponding to approximately 2K clips. We use only the left view of the first exterior camera (\texttt{exterior\_image\_1\_left}) for all experiments. Videos are processed at 15 FPS and segmented into 37-frame clips with a resolution of $480\times704$. All compared methods are trained and evaluated on the same subset, camera view, and video preprocessing configuration.

\section{Training and Evaluation Protocols}
\label{app:training}
\label{app:eval_protocol}

\subsection{Training Configuration}

The state predictor is initialized from Qwen3-VL-2B-Instruct, and the video generator is initialized from Wan2.1-T2V-1.3B or Wan2.2-TI2V-5B according to the evaluated model configuration. The Qwen3-VL backbone is frozen, while the learnable world and ego queries and the SR-MoE video generator are optimized during training.

We use $N_w=192$ world queries and $N_e=64$ ego queries. For both Wan backbones, the first 24 DiT blocks form the shared Preceding Expert, while the final six blocks are duplicated to initialize independently optimized World and Ego Experts.

On HTEWorld, training is performed for four epochs on $16\times$ NVIDIA A100 80GB GPUs with a global batch size of 64 and a learning rate of $1\times10^{-5}$. We maintain an exponential moving average of the model weights with decay 0.99 and use the averaged checkpoint for evaluation. Unless otherwise specified, the additional analyses in this appendix use WEM with the Wan2.2 backbone.

PAN uses the same Qwen3-VL and Wan initialization but predicts a unified state and retains a single generation stream. All compared video generators are fine-tuned rather than adapted only through LoRA. On DROID, the compared methods follow the same optimization protocol on the trajectory-disjoint subset described in Appendix~\ref{app:external_annotation}.

\subsection{Multi-Turn Generation and Baseline Adaptation}

WEM generates video chunks autoregressively. The first round is conditioned on the initial observation and the current instruction. For subsequent rounds, the state predictor receives the initial observation, previous instruction-video pairs, and the current instruction. The visual history consists of previously generated videos rather than reference future frames.

Cosmos-Predict2.5 and WoW-7B require adaptation to this multi-turn setting. During training, both models use a mixed conditioning strategy: with probability 0.9, a clip is conditioned on the preceding chunk, and with probability 0.1, it is initialized from the first frame. This exposes the models to both trajectory continuation and trajectory initialization during training.

At evaluation, Cosmos-Predict2.5 uses \texttt{image2world} for the first generation round and \texttt{video2world} for subsequent rounds. Later rounds are conditioned on the last ten latent frames of the preceding generated chunk.

WoW-7B uses \texttt{video2world} throughout generation. For the first round, the conditioning video is constructed by repeating the initial frame 41 times. For subsequent rounds, a 41-frame tail from the preceding generated video is used. WoW generates 82 frames internally and discards the 41 conditioning frames. The remaining 41 generated frames are retained as the conditioning video for the next round and uniformly subsampled to 37 frames for evaluation.

On HTEWorld, all methods are evaluated using 37 output frames per chunk at $480\times480$ resolution and 16 FPS. We use 35 diffusion steps. The guidance scale is set to 5.0 for all methods.

\subsection{Evaluation Protocol}

We adopt the video quality evaluation protocol introduced by WorldArena~\citep{bib:worldarena}, which comprises 16 metrics across six dimensions: visual quality, motion quality, content consistency, physics adherence, 3D accuracy, and controllability. We report the aggregated EWMScore as the primary metric.
The six HTEWorld-specific diagnostics introduced in \autoref{app:metrics} evaluate complementary properties of multi-turn generation, including cross-chunk continuity, state preservation, and navigation-manipulation transitions.

On WorldArena, WEM is evaluated following the benchmark protocol. The other results reported in the main-text WorldArena comparison are taken from the original WorldArena paper. The additional Mask2IV comparison in \autoref{app:mask2iv} is evaluated by us. On DROID, Wan2.2, PAN, and WEM are trained and evaluated on the same trajectory-disjoint subset described in \autoref{app:external_annotation}.

\section{Alternative Designs}
\label{app:variants}

\input{Figures/fig2_views}

This section provides the implementation details of the variants used in the SR-MoE ablation study in Sec.~\ref{subsec:ablations}. The ablation examines two architectural design choices: \emph{Token Separation}, which determines whether world and ego tokens are separated according to the predicted assignment before subsequent expert processing, and \emph{Expert Specialization}, which determines whether separate World and Ego Experts are used. We additionally compare two choices of \emph{Assignment Proxy} used to derive the world-ego assignment: a semantic ego mask identifying the robot and currently manipulated objects, and a motion proxy capturing motion induced by the robot and manipulated objects after camera-motion compensation. The two assignment views are illustrated in \autoref{fig:views}. We use \emph{Proxy Head} to collectively refer to the Semantic Head and Flow Head used in these two assignment variants. This allows us to examine whether the effectiveness of SR-MoE depends on the choice of assignment proxy. The corresponding generator variants are illustrated in \autoref{fig:framework}.

\input{Figures/fig3_framework}

\subsection{Motion-Based World-Ego Assignment}
\label{app:motion_variant}

The default SR-MoE derives the world-ego assignment from a semantic proxy, namely the ego probability mask predicted by the Semantic Head. To evaluate the choice of assignment proxy, we construct a motion-based alternative that instead predicts a motion proxy corresponding to motion induced by the robot and manipulated objects.

The motion-based variant replaces the Semantic Head with a Flow Head while keeping the remaining generator structure unchanged. The Flow Head takes intermediate features from four selected Preceding Expert blocks, fuses them using learned per-block weights, and decodes the fused representation with a multi-scale U-Net head to predict a two-dimensional flow vector
\begin{equation}
\widehat{\mathbf{F}}^{\mathrm{obj}}_{k,t,i}
\in
\mathbb{R}^{2},
\end{equation}
where $\widehat{\mathbf{F}}^{\mathrm{obj}}_{k,t,i}$ denotes the predicted residual flow at token $(t,i)$ in generation round $k$.
We convert the predicted flow vector into a scalar motion magnitude
\begin{equation}
\widehat{f}_{k,t,i}
=
\left\|
\widehat{\mathbf{F}}^{\mathrm{obj}}_{k,t,i}
\right\|_2.
\end{equation}
Given a motion threshold $\tau_{\mathrm{flow}}$, the corresponding ego and world token sets are defined as
\begin{equation}
\mathcal{I}^e_k
=
\{(t,i)\mid\widehat{f}_{k,t,i}\geq\tau_{\mathrm{flow}}\},
\qquad
\mathcal{I}^w_k
=
\{(t,i)\mid\widehat{f}_{k,t,i}<\tau_{\mathrm{flow}}\}.
\end{equation}

For variants with Token Separation, we apply the same Neighbor-Expanded Assignment as in the default SR-MoE to obtain
$\widetilde{\mathcal{I}}^w_k$ and
$\widetilde{\mathcal{I}}^e_k$ before subsequent expert processing.
For variants without Token Separation, the motion proxy is used only to combine the outputs of the World and Ego Experts after expert processing. We therefore convert the predicted motion magnitude into a continuous world weight
\begin{equation}
\alpha_{k,t,i}
=
\sigma\!\left(
\frac{
\tau_{\mathrm{flow}}
-
\widehat{f}_{k,t,i}
}{
\delta_{\mathrm{flow}}
}
\right),
\end{equation}
where $\delta_{\mathrm{flow}}$ is a constant controlling the transition sharpness. Small residual motion places greater weight on the World Expert, while large residual motion places greater weight on the Ego Expert.
The expert outputs are combined as
\begin{equation}
\widehat{\mathbf{Z}}_k
=
\alpha_k\odot\widehat{\mathbf{Z}}^w_k
+
(1-\alpha_k)\odot\widehat{\mathbf{Z}}^e_k,
\end{equation}
where $\alpha_k$ collects the token-wise weights $\alpha_{k,t,i}$, $\odot$ denotes element-wise multiplication, and $\alpha_k$ is broadcast along the feature dimension to match the expert outputs.

The semantic- and motion-based variants therefore differ only in the proxy used to derive the world-ego assignment: an ego probability mask in the semantic variant and robot- and manipulated-object-induced residual flow in the motion-based variant.

\subsection{SR-MoE Ablation Variants}
\label{app:sr_moe_variants}

The three generator variants shown in Figure~\ref{fig:framework} independently control Token Separation and Expert Specialization while sharing the same state predictor and Preceding Expert.

\paragraph{Token Separation only.}
This variant enables Token Separation while removing Expert Specialization. Starting from the Preceding Expert output $\mathbf{Z}^P_k$, the proxy-derived world and ego assignments are expanded using Neighbor-Expanded Assignment, and the corresponding token groups are processed by the same shared expert.
Although both groups share the same expert parameters, they are processed separately within the expert. Tokens belonging exclusively to the world group are not visible to the ego group, and tokens belonging exclusively to the ego group are not visible to the world group. Tokens in the overlapping region introduced by Neighbor-Expanded Assignment are included in both groups. In addition, the world group is conditioned only on $\mathbf{S}^w_k$, while the ego group is conditioned only on $\mathbf{S}^e_k$.
The two outputs are then restored to their original spatiotemporal positions, with overlapping positions averaged. This variant therefore isolates assignment-based token separation while keeping the generation parameters shared between the world and ego.

\paragraph{Expert Specialization only.}
This variant removes Token Separation while retaining Expert Specialization. The complete Preceding Expert output $\mathbf{Z}^P_k$ is sent to independently optimized World and Ego Experts, each conditioned on its corresponding state.
The proxy-derived assignment is applied only after expert processing to combine the two outputs. For semantic assignment, the predicted ego probability mask $\mathbf{M}_k$ provides the fusion weights:
\begin{equation}
\widehat{\mathbf{Z}}_k
=
(1-\mathbf{M}_k)\odot\widehat{\mathbf{Z}}^w_k
+
\mathbf{M}_k\odot\widehat{\mathbf{Z}}^e_k,
\end{equation}
where $\mathbf{M}_k$ is broadcast along the feature dimension to match the expert outputs.
For motion-based assignment, the flow-derived weights $\alpha_k$ defined in Appendix~\ref{app:motion_variant} are used instead.
This variant therefore isolates the effect of specialized World and Ego parameters without restricting either expert to a role-specific token group.

\paragraph{Token Separation + Expert Specialization.}
The proxy-derived assignment is expanded by Neighbor-Expanded Assignment, and the resulting world and ego token groups are routed to independently optimized World and Ego Experts, each conditioned only on its corresponding state.
The expert outputs are subsequently restored to their original spatiotemporal positions. Tokens covered by only one expanded assignment use the corresponding expert output directly, while tokens in the overlapping region are processed by both experts and their outputs are averaged.
This variant combines assignment-based token separation with specialized generation parameters. Together with the two variants above, it allows the contributions of Token Separation and Expert Specialization to be evaluated independently under the same overall framework.

\section{HTEWorld-Specific Metrics and Evaluation}
\label{app:metrics}

\subsection{Metric Definitions}

The WorldArena metrics remain our primary evaluation protocol. HTEWorld additionally evaluates continuous generation across multiple instruction rounds containing both navigation and manipulation. We therefore introduce six complementary diagnostics. RCBD, LPSA, and CISR examine multi-turn continuity and state evolution, while PMPA, CPDM, and FPHS focus on navigation-manipulation behavior and transitions between the two phases.

Let
$\mathbf{V}=(\mathbf{V}_1,\ldots,\mathbf{V}_K)$
denote a generated trajectory and
$\mathbf{V}^{*}=(\mathbf{V}^{*}_1,\ldots,\mathbf{V}^{*}_K)$
its reference, with chunks aligned by instruction round. Each chunk has a phase label
$p_k\in\{\mathrm{Nav},\mathrm{Manip}\}$.
The first and last frames of $\mathbf{V}_k$ are
$\mathbf{V}_k^{(1)}$ and $\mathbf{V}_k^{(-1)}$.
Let $d_p$ denote LPIPS distance~\citep{bib:lpips} and
$\phi_{\mathrm{CLIP}}(\cdot)$ the CLIP-based visual representation~\citep{bib:clip} of a frame set or video chunk. Feature comparisons use cosine similarity. Each metric is computed per trajectory and then averaged over the evaluation set.

\paragraph{Rollout Chunk-Boundary Dynamics (RCBD).}
RCBD evaluates whether the appearance and motion changes across adjacent generated chunks match those of the reference trajectory. For each adjacent pair, we compute the appearance gap
\begin{equation}
b_k
=
d_p
\left(
\mathbf{V}_k^{(-1)},
\mathbf{V}_{k+1}^{(1)}
\right).
\end{equation}

To characterize the corresponding motion change, let
$\mathbf{F}_k^{-}$ denote the optical flow between the last two frames of $\mathbf{V}_k$, and
$\mathbf{F}_{k+1}^{+}$ the optical flow between the first two frames of $\mathbf{V}_{k+1}$. We define the motion gap as
\begin{equation}
m_k
=
\frac{1}{|\Omega|}
\sum_{\mathbf{p}\in\Omega}
\left\|
\mathbf{F}_k^{-}(\mathbf{p})
-
\mathbf{F}_{k+1}^{+}(\mathbf{p})
\right\|_2,
\end{equation}
where $\Omega$ denotes the image domain. Let $b_k^*$ and $m_k^*$ denote the corresponding appearance and motion gaps computed from the reference trajectory.

We compare the generated and reference gaps using the symmetric ratio score
\begin{equation}
\mathcal{S}(x,y)
=
\exp\left(
-\left|
\log
\frac{x+\epsilon}{y+\epsilon}
\right|
\right),
\end{equation}
where $\epsilon$ is a small constant for numerical stability. RCBD is then defined as
\begin{equation}
\mathrm{RCBD}
=
\frac{1}{K-1}
\sum_{k=1}^{K-1}
\sqrt{
\mathcal{S}(b_k,b_k^*)
\mathcal{S}(m_k,m_k^*)
}.
\end{equation}
Matching reference boundary dynamics avoids rewarding frozen or overly smooth transitions.

\paragraph{Late-Prefix State Alignment (LPSA).}
LPSA measures whether the generated trajectory reaches visual states similar to the reference trajectory after each instruction round, with greater weight placed on later rounds. Let
$\mathbf{V}_k^{[-W:]}$
and
$\mathbf{V}_k^{*[-W:]}$
denote the last $W=4$ frames of the generated and reference chunks, respectively. Their state similarity is
\begin{equation}
r_k
=
\operatorname{cos}
\left(
\phi_{\mathrm{CLIP}}(\mathbf{V}_k^{[-W:]}),
\phi_{\mathrm{CLIP}}(\mathbf{V}_k^{*[-W:]})
\right),
\end{equation}
where $\operatorname{cos}(\cdot,\cdot)$ denotes cosine similarity. We compute
\begin{equation}
\mathrm{LPSA}
=
\frac{
\sum_{k=1}^{K} k\,r_k
}{
\sum_{k=1}^{K} k
}.
\end{equation}
The weighting emphasizes later states, where errors accumulated over previous rounds become increasingly important.

\paragraph{Chunk Instruction-Step Retrieval (CISR).}
CISR evaluates whether each generated chunk corresponds to the correct stage of the reference trajectory.
For each generated chunk $\mathbf{V}_k$, we rank all reference chunks
$\{\mathbf{V}_j^*\}_{j=1}^{K}$
from the same trajectory by
$\operatorname{cos}\!\left(
\phi_{\mathrm{CLIP}}(\mathbf{V}_k),
\phi_{\mathrm{CLIP}}(\mathbf{V}_j^*)
\right)$.
Let $\rho_k$ denote the rank of the corresponding reference chunk $\mathbf{V}_k^*$, where rank 1 indicates the highest similarity. We define
\begin{equation}
\mathrm{CISR}
=
\frac{1}{K}
\sum_{k=1}^{K}
\frac{1}{\rho_k}.
\end{equation}
A higher CISR indicates better alignment with the correct instruction step along the reference trajectory.

\paragraph{Phase-Matched Motion Profile Alignment (PMPA).}
PMPA evaluates whether the generated trajectory exhibits motion patterns similar to the reference within the same instruction phase. Since navigation and manipulation differ substantially in their characteristic motion, we compare phase-matched generated and reference chunks.

For each consecutive frame pair, we estimate optical flow using RAFT~\citep{bib:raft}. Let $u$ denote the resulting flow-magnitude field. We construct a four-dimensional motion descriptor
\begin{equation}
\mathbf{q}
=
\left[
\bar{u}/D,\;
\widetilde{u}/D,\;
\log\left(1+\frac{\widetilde{u}}{\bar{u}+\epsilon}\right),\;
\mathcal{E}(u)
\right],
\end{equation}
where $\bar{u}$ is the median flow magnitude, $\widetilde{u}$ is the mean of the largest 20\% of flow magnitudes, $D$ is the frame diagonal, $\mathcal{E}(u)$ denotes the normalized entropy of the flow-magnitude distribution, and $\epsilon$ is a small constant for numerical stability.

The generated and reference motion-profile sequences are temporally resampled to 16 positions. Let $\mathbf{q}_{k,s}$ and $\mathbf{q}_{k,s}^{*}$ denote the generated and reference descriptors at resampled position $s$ of chunk $k$, respectively. Their discrepancy is
\begin{equation}
\delta_k
=
\frac{1}{16}
\sum_{s=1}^{16}
\left\|
\mathbf{q}_{k,s}
-
\mathbf{q}_{k,s}^{*}
\right\|_2.
\end{equation}
We define the PMPA score of chunk $k$ as
\begin{equation}
s_k^{\mathrm{PMPA}}
=
\exp
\left(
-\delta_k/\tau_{\mathrm{PMPA}}
\right),
\end{equation}
where $\tau_{\mathrm{PMPA}}$ controls the sensitivity to motion-profile discrepancies. The trajectory-level PMPA is then
\begin{equation}
\mathrm{PMPA}
=
\frac{1}{K}
\sum_{k=1}^{K}
s_k^{\mathrm{PMPA}},
\end{equation}
where $K$ is the number of evaluated chunks. Each generated chunk is compared with its corresponding reference chunk from the same navigation or manipulation phase.

\paragraph{Cross-Phase Discriminative Margin (CPDM).}
CPDM evaluates whether a generated chunk is more similar to its corresponding reference than to reference chunks from the opposite phase. For chunk $k$, we define the positive similarity as
\begin{equation}
r_k^+
=
\operatorname{cos}
\left(
\phi_{\mathrm{CLIP}}(\mathbf{V}_k),
\phi_{\mathrm{CLIP}}(\mathbf{V}_k^*)
\right),
\end{equation}
and the strongest opposite-phase similarity as
\begin{equation}
r_k^-
=
\max_{j:\,p_j\neq p_k}
\operatorname{cos}
\left(
\phi_{\mathrm{CLIP}}(\mathbf{V}_k),
\phi_{\mathrm{CLIP}}(\mathbf{V}_j^*)
\right).
\end{equation}
We compute
\begin{equation}
\mathrm{CPDM}
=
\frac{1}{K}
\sum_{k=1}^{K}
\sigma
\left(
\frac{
r_k^+-r_k^-
}{
\tau_{\mathrm{CPDM}}
}
\right),
\qquad
\tau_{\mathrm{CPDM}}=0.05,
\end{equation}
where $\sigma$ denotes the sigmoid function. A higher score indicates that the generated trajectory better preserves the distinction between navigation and manipulation phases.

\paragraph{Frontier Phase-Hop State Consistency (FPHS).}
FPHS focuses specifically on transitions between navigation and manipulation. For each transition boundary satisfying
$p_k\neq p_{k+1}$,
we extract $R=4$ frames on both sides of the boundary from both the generated and reference trajectories.

We compute optical-flow magnitudes over the reference transition window and accumulate them temporally to obtain a spatial change map. The top-20\% locations in this map define the region with the strongest reference transition dynamics. The generated and reference transition windows are restricted to this region and encoded using
$\phi_{\mathrm{CLIP}}(\cdot)$.
Their cosine similarity provides the score for the corresponding transition boundary. FPHS is obtained by averaging these scores over all navigation-manipulation transition boundaries in the trajectory.

\subsection{Human Correlation and Interpretation}
\label{app:human_evaluation}


We evaluate the agreement between the six HTEWorld-specific diagnostics and human judgments using 300 generated trajectories from each of seven systems, yielding 2,100 complete multi-turn rollouts. We recruit 15 volunteers, each of whom independently evaluates all 2,100 anonymized rollouts on a five-point scale along two dimensions. \emph{Perceived task completion} measures whether the generated video appears to accomplish the instructed behavior, while \emph{visual realism} evaluates visual coherence, temporal consistency, and physical plausibility. For each rollout, ratings are averaged across the 15 annotators, and Pearson correlation is computed between the resulting human scores and each diagnostic metric.

\begin{table}[t]
\centering
\small
\caption{Pearson correlation between HTEWorld-specific metrics and human ratings of complete multi-turn rollouts. Task completion is judged from generated videos rather than measured through robot execution.}
\label{tab:app_human_correlation}
\begin{tabular}{lrr}
\toprule
Metric & Perceived task completion & Visual realism \\
\midrule
RCBD & \textbf{0.726} & 0.584 \\
LPSA & 0.710 & 0.649 \\
CISR & 0.569 & 0.706 \\
PMPA & 0.651 & 0.622 \\
CPDM & 0.465 & 0.653 \\
FPHS & 0.516 & \textbf{0.834} \\
\bottomrule
\end{tabular}
\end{table}

As shown in \autoref{tab:app_human_correlation}, all six metrics are positively correlated with both rating dimensions. RCBD shows the strongest correlation with perceived task completion, while FPHS shows the strongest correlation with visual realism. LPSA also correlates strongly with both dimensions. The different correlation patterns indicate that the six metrics capture complementary properties of multi-turn generation rather than serving as interchangeable measures.

The metrics nevertheless remain diagnostic rather than direct measures of robot task success. RCBD and PMPA depend on optical-flow estimates and cannot establish that the correct object is manipulated. LPSA, CISR, and CPDM rely on visual representations that may overlook fine-grained contact, geometry, or physical violations. CISR also depends on the number and visual similarity of reference chunks within a trajectory. FPHS focuses on reference-derived transition regions and may miss errors elsewhere in the scene. More generally, reference-based evaluation may penalize a plausible rollout that differs from the recorded trajectory. We therefore use these diagnostics together with the primary WorldArena metrics and qualitative evaluation.

\subsection{Results on HTEWorld-Specific Metrics}
\label{app:metrics_results}

\autoref{tab:hteworld_specific_metrics} reports results on the six HTEWorld-specific metrics. PAN and WEM use the Wan2.2 backbone in this comparison. WEM achieves the best result on all six metrics. In particular, RCBD increases from 0.27 to 0.31, CISR from 0.49 to 0.57, PMPA from 0.50 to 0.54, and CPDM from 0.46 to 0.52 compared with PAN. LPSA and FPHS also improve from 0.86 to 0.87 and from 0.88 to 0.89, respectively. These results provide complementary evidence for improved multi-turn continuity and navigation-manipulation generation beyond the aggregate EWMScore reported in the main text.

\begin{table}[t]
\centering
\small
\setlength{\tabcolsep}{5pt}
\caption{Results on the six HTEWorld-specific metrics. Higher is better. PAN and WEM use the Wan2.2 backbone.}
\label{tab:hteworld_specific_metrics}
\begin{tabular}{lcccccc}
\toprule
Model & RCBD & LPSA & CISR & PMPA & CPDM & FPHS \\
\midrule
WoW-7B & 0.23 & 0.83 & 0.49 & 0.45 & 0.47 & 0.85 \\
Cosmos-Predict2.5-2B & 0.24 & 0.83 & 0.50 & 0.47 & 0.48 & 0.86 \\
Cosmos-Predict2.5-14B & 0.26 & 0.83 & 0.51 & 0.48 & 0.49 & 0.85 \\
PAN & 0.27 & 0.86 & 0.49 & 0.50 & 0.46 & 0.88 \\
\rowcolor[HTML]{D9E1F2}
WEM & \textbf{0.31} & \textbf{0.87} & \textbf{0.57} & \textbf{0.54} & \textbf{0.52} & \textbf{0.89} \\
\bottomrule
\end{tabular}
\end{table}

\section{Additional Experiments and Analyses}
\label{app:additional_results}

\subsection{World-Ego Assignment and Expert Specialization}
\label{app:qualitative}

\paragraph{Assignment accuracy.}
We evaluate the world-ego masks predicted by the Semantic Head on the HTEWorld validation set using the default WEM configuration. At the final denoising step, the predicted masks achieve an Ego IoU of 70.3\%, a World IoU of 91.7\%, and a mean IoU of 81.0\%. The predictions are already accurate at an early denoising stage, such as the second step, and improve only moderately during later steps. The lower Ego IoU reflects the greater difficulty of identifying the smaller robot-object interaction regions compared with the larger world region.

\paragraph{Dynamic role transitions.}
The semantic world-ego assignment changes with the current interaction rather than assigning a fixed role to each object category. An object belongs to the world before manipulation, becomes part of the ego while it is manipulated or carried, and can return to the world after the interaction. For example, in a cleanup trajectory, a can is initially part of the surrounding scene and is reassigned to the ego after the robot picks it up. The assignment therefore follows the interaction state of an object rather than whether its pixels are moving.

\paragraph{Role expert specialization.}
\autoref{fig:role_expert_visualization} visualizes the regions associated with the World and Ego Experts together with the recomposed WEM output. The Ego Expert view contains the robot and currently manipulated objects, while the World Expert view contains the remaining scene. The assignments also evolve as objects enter or leave the interaction. Since these views are obtained by masking the expert outputs according to the predicted assignment, they visualize which regions are associated with each expert rather than independently proving expert specialization. The quantitative generator ablations in the main text provide the corresponding evidence for the benefit of specialized generation parameters.

\input{Figures/role_expert_visualization}

\subsection{Comparison with Mask2IV}
\label{app:mask2iv}

Mask2IV~\citep{bib:mask2iv} uses object masks to guide interaction-centric video generation. We compare WEM with its Text-conditioned Trajectory Generation (TT-Gen) variant on WorldArena. Mask2IV is fine-tuned on the same WorldArena training split as WEM, with both methods using masks from the same SAM~3 annotation pipeline.

The two methods differ in how masks are used during generation. Mask2IV takes a SAM~3-derived initial object mask as an additional inference condition and predicts a mask trajectory to guide a shared video generator. WEM requires no mask at inference; instead, it predicts the world-ego assignment during denoising and routes tokens to separate World and Ego Experts. Because the methods differ in both architecture and inference conditions, this serves as an additional benchmark comparison rather than a controlled isolation of token routing; the controlled SR-MoE ablations are provided in Sec.~\ref{subsec:ablations}.

\begin{table}[t]
\centering
\caption{Additional comparison with Mask2IV on WorldArena, evaluated by us. Both methods use text instructions; Mask2IV additionally receives an initial object mask at inference.}
\label{tab:app_mask2iv}
\begin{tabular}{lc}
\toprule
Method & EWMScore $\uparrow$ \\
\midrule
Mask2IV TT-Gen & 52.36 \\
\rowcolor[HTML]{D9E1F2}
WEM (Wan2.2) & \textbf{58.10} \\
\bottomrule
\end{tabular}
\end{table}

As shown in \autoref{tab:app_mask2iv}, WEM achieves an EWMScore of 58.10, outperforming Mask2IV TT-Gen by 5.74 points. Since WorldArena focuses on manipulation, this result further shows that the advantage of WEM is not limited to trajectories containing navigation.

\section{Limitations and Future Work}
\label{app:limitations}

\paragraph{Long-horizon error accumulation.}
WEM improves long-horizon scene consistency and instruction following, but autoregressive generation still accumulates errors over successive rounds. Incorrect object states or scene geometry in one generated chunk can affect subsequent predictions, and information from early rounds may gradually be lost as the trajectory becomes longer. The current world and ego states summarize the available context but do not provide persistent memory over arbitrarily long horizons. Incorporating more explicit long-term memory or hierarchical representations of task progress is a promising direction.

\paragraph{Real-world hybrid evaluation.}
DROID allows us to evaluate WEM on real-world tabletop manipulation videos, but it does not contain long-horizon trajectories with interleaved navigation and manipulation. The lack of corresponding real-world data therefore limits evaluation of the complete hybrid setting outside simulation. Extending the evaluation to real trajectories combining large viewpoint changes, navigation, manipulation, sensing noise, and diverse physical interactions would provide a more comprehensive test of long-horizon embodied generation.

\paragraph{Downstream scope.}
Our experiments evaluate embodied video generation rather than closed-loop robot control. Improvements in EWMScore, HTEWorld-specific diagnostics, and perceived task completion therefore do not directly establish improved planning or policy performance. Evaluating whether the generated world and ego evolution can benefit downstream planning, policy learning, or real-robot control remains an important direction for future work.

%% file: Figures/fig2_views.tex
\begin{figure}[t!]
\begin{center}
\includegraphics[width=0.95\linewidth]{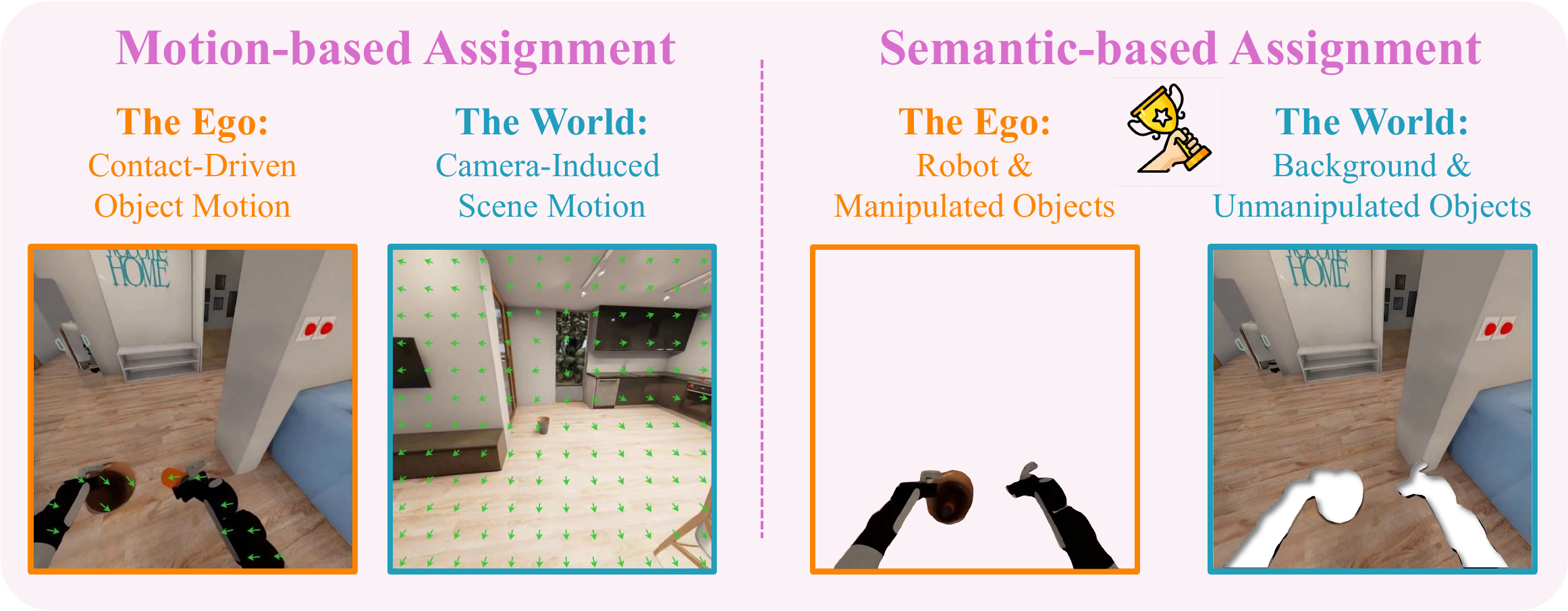}
\caption{
Comparison of motion-based and semantic-based world-ego assignments.
The motion-based assignment derives the world-ego assignment from a motion proxy, where the ego corresponds to residual motion induced by the robot and manipulated objects after camera-motion compensation, while the world corresponds to camera-induced background motion.
The semantic-based assignment derives the world-ego assignment from a semantic ego mask, where the ego consists of the robot and currently manipulated objects, while the world consists of the background and unmanipulated objects.
}
\label{fig:views}
\end{center}
\end{figure}

%% file: Figures/fig3_framework.tex
\begin{figure}[t!]
\begin{center}
\includegraphics[width=0.95\linewidth]{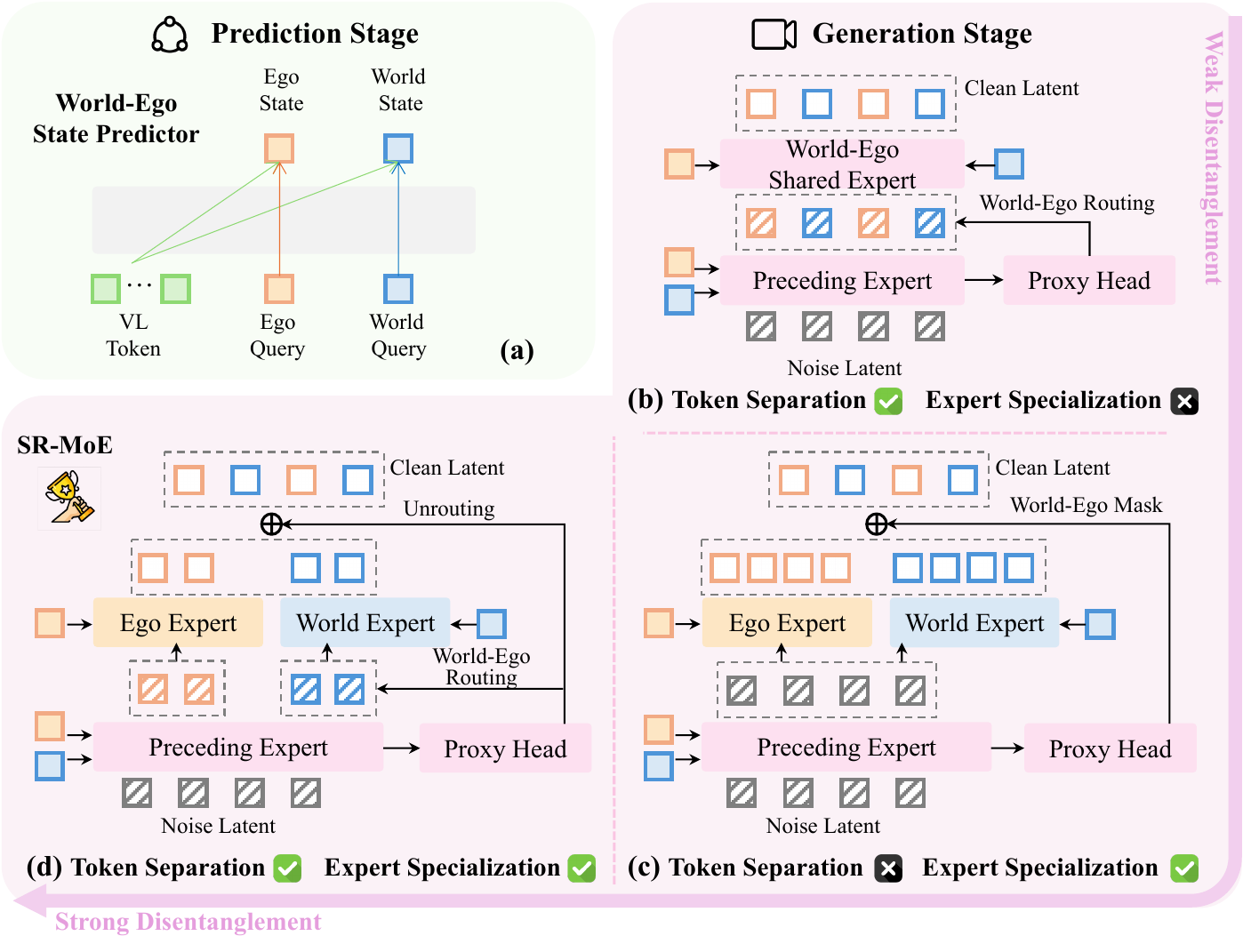}
\caption{
Architectural variants of SR-MoE evaluated in the ablation study.
Our framework contains a shared prediction stage and three generator variants with different combinations of Token Separation and Expert Specialization. (a) The prediction stage uses world and ego queries to extract two role-specific states from vision-language tokens. A Proxy Head predicts a world-ego proxy from the Preceding Expert features, instantiated as either a semantic ego mask or motion induced by the robot and manipulated objects. The world-ego assignment is then derived from this proxy. (b) With Token Separation only, the assignment separates the Preceding Expert output into world and ego token groups, which are processed independently by the same shared expert under their corresponding states. (c) With Expert Specialization only, the complete token sequence is processed by separate World and Ego Experts, and the proxy-derived assignment is used to combine their outputs. (d) The full SR-MoE combines both designs by routing the assigned world and ego token groups to their corresponding specialized experts and restoring the processed tokens to their original spatiotemporal positions.
}
\label{fig:framework}
\end{center}
\end{figure}

%% file: Figures/role_expert_visualization.tex
\begin{figure}[t]
\centering
\includegraphics[width=\linewidth]{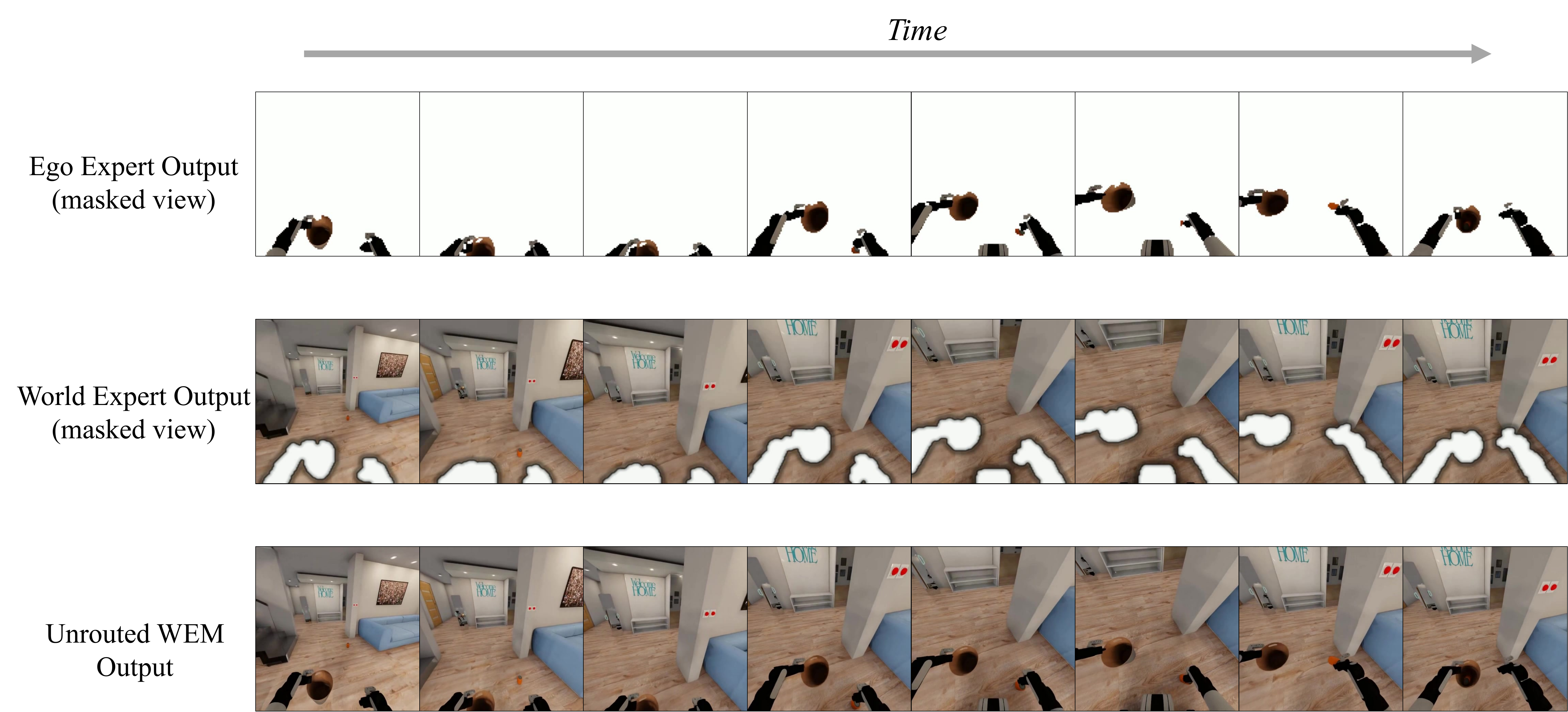}
\caption{Masked views of the role expert outputs and the recomposed WEM output. The Ego Expert output is displayed in ego-assigned regions, and the World Expert output in world-assigned regions, using the assignment predicted by the Semantic Head. Complementary regions are left blank for visualization. The bottom row shows the output after restoring the expert outputs to their original positions. The masking is used for visualization only.}
\label{fig:role_expert_visualization}
\end{figure}

%% file: main.bib
@String(PAMI = {IEEE Trans. Pattern Anal. Mach. Intell.})

@String(CVPR= {IEEE Conf. Comput. Vis. Pattern Recog.})

@String(ICCV= {Int. Conf. Comput. Vis.})

@String(ECCV= {Eur. Conf. Comput. Vis.})

@String(NIPS= {Adv. Neural Inform. Process. Syst.})

@String(ICLR = {Int. Conf. Learn. Represent.})

@String(AAAI = {AAAI})

@String(PAMI  = {IEEE TPAMI})

@String(CVPR  = {CVPR})

@String(ICCV  = {ICCV})

@String(ECCV  = {ECCV})

@String(NIPS  = {NeurIPS})

@String(ICLR  = {ICLR})

@String(ICML = {ICML})

@String(CoRL = {CoRL})

@String(RSS = {RSS})

@String(SIGGRAPH = {SIGGRAPH})

@inproceedings{bib:world_models,
  title = {Recurrent World Models Facilitate Policy Evolution},
  author = {Ha, David and Schmidhuber, J{\"u}rgen},
  _booktitle={Proceedings of the Neural Information Processing Systems (NeurIPS)},
  booktitle=NIPS,
  year = {2018},
}

@article{bib:dreamerv3,
  title = {Mastering Diverse Control Tasks Through World Models},
  author = {Hafner, Danijar and Pasukonis, Jurgis and Ba, Jimmy and Lillicrap, Timothy},
  journal = {Nature},
  year = {2025},
}

@article{bib:cosmos_predict,
  title = {World simulation with video foundation models for physical {AI}},
  author = {Ali, Arslan and Bai, Junjie and Bala, Maciej and Balaji, Yogesh and Blakeman, Aaron and Cai, Tiffany and Cao, Jiaxin and Cao, Tianshi and Cha, Elizabeth and Chao, Yu-Wei and others},
  journal = {arXiv preprint arXiv:2511.00062},
  year = {2025},
}

@article{bib:cosmos,
  title = {Cosmos world foundation model platform for physical ai},
  author = {Agarwal, Niket and Ali, Arslan and Bala, Maciej and Balaji, Yogesh and Barker, Erik and Cai, Tiffany and Chattopadhyay, Prithvijit and Chen, Yongxin and Cui, Yin and Ding, Yifan and others},
  journal = {arXiv preprint arXiv:2501.03575},
  year = {2025},
}

@article{bib:pan,
  title = {{PAN}: A world model for general, interactable, and long-horizon world simulation},
  author = {Xiang, Jiannan and Gu, Yi and Liu, Zihan and Feng, Zeyu and Gao, Qiyue and Hu, Yiyan and Huang, Benhao and Liu, Guangyi and Yang, Yichi and Zhou, Kun and others},
  journal = {arXiv preprint arXiv:2511.09057},
  year = {2025},
}

@inproceedings{bib:unisim,
  title = {Learning Interactive Real-World Simulators},
  author = {Yang, Mengjiao and Du, Yilun and Ghasemipour, Seyed Kamyar Seyed and Tompson, Jonathan and Kaelbling, Leslie Pack and Schuurmans, Dale and Abbeel, Pieter},
  _booktitle={Proceedings of the International Conference on Learning Representations (ICLR)},
  booktitle=ICLR,
  year = {2024},
}

@inproceedings{bib:genie,
  title = {Genie: Generative Interactive Environments},
  author = {Bruce, Jake and Dennis, Michael D and Edwards, Ashley and Parker-Holder, Jack and Shi, Yuge and Hughes, Edward and Lai, Matthew and Mavalankar, Aditi and Steigerwald, Richie and Apps, Chris and others},
  _booktitle={Proceedings of the International Conference on Machine Learning (ICML)},
  booktitle=ICML,
  year = {2024},
}

@inproceedings{bib:robodreamer,
  title = {RoboDreamer: Learning Compositional World Models for Robot Imagination},
  author = {Zhou, Siyuan and Du, Yilun and Chen, Jiaben and Li, Yandong and Yeung, Dit-Yan and Gan, Chuang},
  _booktitle={Proceedings of the International Conference on Machine Learning (ICML)},
  booktitle=ICML,
  year = {2024},
}

@inproceedings{bib:tesseract,
  title = {TesserAct: Learning 4D Embodied World Models},
  author = {Zhen, Haoyu and Sun, Qiao and Zhang, Hongxin and Li, Junyan and Zhou, Siyuan and Du, Yilun and Gan, Chuang},
  _booktitle={Proceedings of the IEEE/CVF International Conference on Computer Vision (ICCV)},
  booktitle=ICCV,
  year = {2025},
}

@inproceedings{bib:gem,
  title = {GEM: A Generalizable Ego-Vision Multimodal World Model for Fine-Grained Ego-Motion, Object Dynamics, and Scene Composition Control},
  author = {Hassan, Mariam and Stapf, Sebastian and Rahimi, Ahmad and Rezende, Pedro and Haghighi, Yasaman and Br{\"u}ggemann, David and Katircioglu, Isinsu and Zhang, Lin and Chen, Xiaoran and Saha, Suman and others},
  _booktitle={Proceedings of the IEEE/CVF Conference on Computer Vision and Pattern Recognition (CVPR)},
  booktitle=CVPR,
  year = {2025},
}

@inproceedings{bib:dreamgen,
  title = {DreamGen: Unlocking Generalization in Robot Learning through Video World Models},
  author = {Jang, Joel and Ye, Seonghyeon and Lin, Zongyu and Xiang, Jiannan and Bjorck, Johan and Fang, Yu and Hu, Fengyuan and Huang, Spencer and Kundalia, Kaushil and Lin, Yen-Chen and others},
  _booktitle={Proceedings of the Conference on Robot Learning (CoRL)},
  booktitle=CoRL,
  year = {2025},
}

@inproceedings{bib:behavior,
  title = {{BEHAVIOR-1K}: A Benchmark for Embodied AI with 1,000 Everyday Activities and Realistic Simulation},
  author = {Li, Chengshu and Zhang, Ruohan and Wong, Josiah and Gokmen, Cem and Srivastava, Sanjana and Mart{\'\i}n-Mart{\'\i}n, Roberto and Wang, Chen and Levine, Gabrael and Lingelbach, Michael and Sun, Jiankai and others},
  _booktitle={Proceedings of the Conference on Robot Learning (CoRL)},
  booktitle=CoRL,
  year = {2023},
}

@inproceedings{bib:worldmodelbench,
  title = {WorldModelBench: Judging Video Generation Models As World Models},
  author = {Li, Dacheng and Fang, Yunhao and Chen, Yukang and Yang, Shuo and Cao, Shiyi and Wong, Justin and Luo, Michael and Wang, Xiaolong and Yin, Hongxu and Gonzalez, Joseph E and others},
  _booktitle={Proceedings of the Neural Information Processing Systems Datasets and Benchmarks Track (NeurIPS Datasets and Benchmarks)},
  booktitle=NIPS,
  year = {2025},
}

@inproceedings{bib:rbench,
  title = {Rethinking Video Generation Model for the Embodied World},
  author = {Deng, Yufan and Pan, Zilin and Zhang, Hongyu and Li, Xiaojie and Hu, Ruoqing and Ding, Yufei and Zou, Yiming and Zeng, Yan and Zhou, Daquan},
  _booktitle={Proceedings of the International Conference on Machine Learning (ICML)},
  booktitle=ICML,
  year = {2026},
}

@article{bib:worldarena,
  title = {WorldArena: A Unified Benchmark for Evaluating Perception and Functional Utility of Embodied World Models},
  author = {Shang, Yu and Li, Zhuohang and Ma, Yiding and Su, Weikang and Jin, Xin and Wang, Ziyou and Jin, Lei and Zhang, Xin and Tang, Yinzhou and Su, Haisheng and others},
  journal = {arXiv preprint arXiv:2602.08971},
  year = {2026},
}

@inproceedings{bib:daydreamer,
  title = {DayDreamer: World Models for Physical Robot Learning},
  author = {Wu, Philipp and Escontrela, Alejandro and Hafner, Danijar and Abbeel, Pieter and Goldberg, Ken},
  _booktitle={Proceedings of the Conference on Robot Learning (CoRL)},
  booktitle=CoRL,
  year = {2023},
}

@article{bib:longscape,
  title = {LongScape: Advancing Long-Horizon Embodied World Models with Context-Aware MoE},
  author = {Shang, Yu and Jin, Lei and Ma, Yiding and Zhang, Xin and Gao, Chen and Wu, Wei and Li, Yong},
  journal = {arXiv preprint arXiv:2509.21790},
  year = {2025},
}

@article{bib:magi,
  title = {Magi-1: Autoregressive video generation at scale},
  author = {Teng, Hansi and Jia, Hongyu and Sun, Lei and Li, Lingzhi and Li, Maolin and Tang, Mingqiu and Han, Shuai and Zhang, Tianning and Zhang, WQ and Luo, Weifeng and others},
  journal = {arXiv preprint arXiv:2505.13211},
  year = {2025},
}

@article{bib:qwen3-vl,
  title = {Qwen3-vl technical report},
  author = {Bai, Shuai and Cai, Yuxuan and Chen, Ruizhe and Chen, Keqin and Chen, Xionghui and Cheng, Zesen and Deng, Lianghao and Ding, Wei and Gao, Chang and Ge, Chunjiang and others},
  journal = {arXiv preprint arXiv:2511.21631},
  year = {2025},
}

@article{bib:wan,
  title = {Wan: Open and advanced large-scale video generative models},
  author = {Wan, Team and Wang, Ang and Ai, Baole and Wen, Bin and Mao, Chaojie and Xie, Chen-Wei and Chen, Di and Yu, Feiwu and Zhao, Haiming and Yang, Jianxiao and others},
  journal = {arXiv preprint arXiv:2503.20314},
  year = {2025},
}

@inproceedings{bib:dpt,
  title = {Vision Transformers for Dense Prediction},
  author = {Ranftl, Ren{\'e} and Bochkovskiy, Alexey and Koltun, Vladlen},
  _booktitle={Proceedings of the IEEE/CVF International Conference on Computer Vision (ICCV)},
  booktitle=ICCV,
  year = {2021},
}

@inproceedings{bib:dreamer,
  title = {Dream to Control: Learning Behaviors by Latent Imagination},
  author = {Hafner, Danijar and Lillicrap, Timothy and Ba, Jimmy and Norouzi, Mohammad},
  _booktitle={Proceedings of the International Conference on Learning Representations (ICLR)},
  booktitle=ICLR,
  year = {2020},
}

@inproceedings{bib:ddpm,
  title = {Denoising Diffusion Probabilistic Models},
  author = {Ho, Jonathan and Jain, Ajay and Abbeel, Pieter},
  _booktitle={Proceedings of the Neural Information Processing Systems (NeurIPS)},
  booktitle=NIPS,
  year = {2020},
}

@inproceedings{bib:image_ldm,
  title = {High-Resolution Image Synthesis with Latent Diffusion Models},
  author = {Rombach, Robin and Blattmann, Andreas and Lorenz, Dominik and Esser, Patrick and Ommer, Bj{\"o}rn},
  _booktitle={Proceedings of the IEEE/CVF Conference on Computer Vision and Pattern Recognition (CVPR)},
  booktitle=CVPR,
  year = {2022},
}

@article{bib:sora,
  title = {Sora: {{A}} Review on Background, Technology, Limitations, and Opportunities of Large Vision Models},
  author = {Liu, Yixin and Zhang, Kai and Li, Yuan and Yan, Zhiling and Gao, Chujie and Chen, Ruoxi and Yuan, Zhengqing and Huang, Yue and Sun, Hanchi and Gao, Jianfeng and He, Lifang and Sun, Lichao},
  journal = {arXiv preprint arXiv:2402.17177},
  year = {2024},
}

@article{bib:the_matrix,
  title = {The Matrix: {{Infinite-horizon}} World Generation with Real-Time Moving Control},
  author = {Feng, Ruili and Zhang, Han and Yang, Zhantao and Xiao, Jie and Shu, Zhilei and Liu, Zhiheng and Zheng, Andy and Huang, Yukun and Liu, Yu and Zhang, Hongyang},
  journal = {arXiv preprint arXiv:2412.03568},
  year = {2024},
}

@article{bib:yume,
  title = {Yume: {{An}} Interactive World Generation Model},
  author = {Mao, Xiaofeng and Lin, Shaoheng and Li, Zhen and Li, Chuanhao and Peng, Wenshuo and He, Tong and Pang, Jiangmiao and Chi, Mingmin and Qiao, Yu and Zhang, Kaipeng},
  journal = {arXiv preprint arXiv:2507.17744},
  year = {2025},
}

@article{bib:live,
  title = {{{LIVE}}: {{Long-horizon}} Interactive Video World Modeling},
  author = {Huang, Junchao and Ye, Ziyang and Hu, Xinting and He, Tianyu and Zhang, Guiyu and Shi, Shaoshuai and Bian, Jiang and Jiang, Li},
  journal = {arXiv preprint arXiv:2602.03747},
  year = {2026},
}

@inproceedings{bib:roboscape,
  title = {RoboScape: Physics-informed Embodied World Model},
  author = {Shang, Yu and Zhang, Xin and Tang, Yinzhou and Jin, Lei and Gao, Chen and Wu, Wei and Li, Yong},
  _booktitle={Proceedings of the Neural Information Processing Systems (NeurIPS)},
  booktitle=NIPS,
  year = {2025},
}

@article{bib:wow,
  title = {{{WoW}}: {{Towards}} a World Omniscient World Model through Embodied Interaction},
  author = {Chi, Xiaowei and Jia, Peidong and Fan, Chun-Kai and Ju, Xiaozhu and Mi, Weishi and Zhang, Kevin and Qin, Zhiyuan and Tian, Wanxin and Ge, Kuangzhi and Li, Hao and Qian, Zezhong and Chen, Anthony and Zhou, Qiang and Jia, Yueru and Liu, Jiaming and Dai, Yong and Wuwu, Qingpo and Bai, Chengyu and Wang, Yu-Kai and Li, Ying and Chen, Lizhang and Bao, Yong and Jiang, Zhiyuan and Zhu, Jiacheng and Tang, Kai and An, Ruichuan and Luo, Yulin and Feng, Qiuxuan and Zhou, Siyuan and Chan, Chi-min and Hou, Chengkai and Xue, Wei and Han, Sirui and Guo, Yike and Zhang, Shanghang and Tang, Jian},
  journal = {arXiv preprint arXiv:2509.22642},
  year = {2025},
}

@inproceedings{bib:ctrl_world,
  title = {Ctrl-World: A Controllable Generative World Model for Robot Manipulation},
  author = {Guo, Yanjiang and Shi, Lucy Xiaoyang and Chen, Jianyu and Finn, Chelsea},
  _booktitle={Proceedings of the International Conference on Learning Representations (ICLR)},
  booktitle=ICLR,
  year = {2026},
}

@inproceedings{bib:raft,
  title = {RAFT: Recurrent All-Pairs Field Transforms for Optical Flow},
  author = {Teed, Zachary and Deng, Jia},
  _booktitle={Proceedings of the European Conference on Computer Vision (ECCV)},
  booktitle=ECCV,
  year = {2020},
}

@inproceedings{bib:lpips,
  title = {The Unreasonable Effectiveness of Deep Features as a Perceptual Metric},
  author = {Zhang, Richard and Isola, Phillip and Efros, Alexei A and Shechtman, Eli and Wang, Oliver},
  _booktitle={Proceedings of the IEEE/CVF Conference on Computer Vision and Pattern Recognition (CVPR)},
  booktitle=CVPR,
  year = {2018},
}

@inproceedings{bib:clip,
  title = {Learning Transferable Visual Models From Natural Language Supervision},
  author = {Radford, Alec and Kim, Jong Wook and Hallacy, Chris and Ramesh, Aditya and Goh, Gabriel and Agarwal, Sandhini and Sastry, Girish and Askell, Amanda and Mishkin, Pamela and Clark, Jack and others},
  _booktitle={Proceedings of the International Conference on Machine Learning (ICML)},
  booktitle=ICML,
  year = {2021},
}

@inproceedings{bib:dit,
  title = {Scalable Diffusion Models with Transformers},
  author = {Peebles, William and Xie, Saining},
  _booktitle={Proceedings of the IEEE/CVF International Conference on Computer Vision (ICCV)},
  booktitle=ICCV,
  year = {2023},
}

@inproceedings{bib:cogvideox,
  title = {{CogVideoX}: Text-to-Video Diffusion Models with an Expert Transformer},
  author = {Yang, Zhuoyi and Teng, Jiayan and Zheng, Wendi and Ding, Ming and Huang, Shiyu and Xu, Jiazheng and Yang, Yuanming and Hong, Wenyi and Zhang, Xiaohan and Feng, Guanyu and others},
  _booktitle={Proceedings of the International Conference on Learning Representations (ICLR)},
  booktitle=ICLR,
  year = {2025},
}

@inproceedings{bib:dice,
  title = {{V-Net}: Fully Convolutional Neural Networks for Volumetric Medical Image Segmentation},
  author = {Milletari, Fausto and Navab, Nassir and Ahmadi, Seyed-Ahmad},
  _booktitle = {Proceedings of the Fourth International Conference on 3D Vision (3DV)},
  booktitle = {3DV},
  _pages = {565--571},
  year = {2016},
}

@inproceedings{bib:flow_matching,
  title = {Flow Matching for Generative Modeling},
  author = {Lipman, Yaron and Chen, Ricky T. Q. and Ben-Hamu, Heli and Nickel, Maximilian and Le, Matt},
  _booktitle={Proceedings of the International Conference on Learning Representations (ICLR)},
  booktitle=ICLR,
  year = {2023},
}

@inproceedings{bib:vid2world,
  title = {{Vid2World}: Crafting Video Diffusion Models to Interactive World Models},
  author = {Huang, Siqiao and Wu, Jialong and Zhou, Qixing and Miao, Shangchen and Long, Mingsheng},
  _booktitle={Proceedings of the International Conference on Learning Representations (ICLR)},
  booktitle=ICLR,
  year = {2026},
}

@misc{bib:genie2,
  title = {Genie 2: A Large-Scale Foundation World Model},
  author = {Parker-Holder, Jack and Ball, Philip and Bruce, Jake and Dasagi, Vibhavari and Holsheimer, Kristian and Kaplanis, Christos and Moufarek, Alexandre and Scully, Guy and Shar, Jeremy and Shi, Jimmy and others},
  year = {2024},
  url = {https://deepmind.google/blog/genie-2-a-large-scale-foundation-world-model/},
  note = {Google DeepMind blog},
}

@article{bib:matrix2,
  title = {Matrix-Game 2.0: An Open-Source, Real-Time, and Streaming Interactive World Model},
  author = {He, Xianglong and Peng, Chunli and Liu, Zexiang and Wang, Boyang and Zhang, Yifan and Cui, Qi and Kang, Fei and Jiang, Biao and An, Mengyin and Ren, Yangyang and others},
  journal = {arXiv preprint arXiv:2508.13009},
  year = {2025},
}

@article{bib:hunyuanvideo,
  title = {{HunyuanVideo}: A Systematic Framework for Large Video Generative Models},
  author = {Kong, Weijie and Tian, Qi and Zhang, Zijian and Min, Rox and Dai, Zuozhuo and Zhou, Jin and Xiong, Jiangfeng and Li, Xin and Wu, Bo and Zhang, Jianwei and others},
  journal = {arXiv preprint arXiv:2412.03603},
  year = {2024},
}

@article{bib:vbench++,
  title = {{VBench++}: Comprehensive and Versatile Benchmark Suite for Video Generative Models},
  author = {Huang, Ziqi and Zhang, Fan and Xu, Xiaojie and He, Yinan and Yu, Jiashuo and Dong, Ziyue and Ma, Qianli and Chanpaisit, Nattapol and Si, Chenyang and Jiang, Yuming and others},
  journal=PAMI,
  year = {2025},
}

@inproceedings{bib:vbench,
  title = {{VBench}: Comprehensive Benchmark Suite for Video Generative Models},
  author = {Huang, Ziqi and He, Yinan and Yu, Jiashuo and Zhang, Fan and Si, Chenyang and Jiang, Yuming and Zhang, Yuanhan and Wu, Tianxing and Jin, Qingyang and Chanpaisit, Nattapol and others},
  _booktitle={Proceedings of the IEEE/CVF Conference on Computer Vision and Pattern Recognition (CVPR)},
  booktitle=CVPR,
  year = {2024},
}

@article{bib:vbench2,
  title = {{VBench}-2.0: Advancing Video Generation Benchmark Suite for Intrinsic Faithfulness},
  author = {Zheng, Dian and Huang, Ziqi and Liu, Hongbo and Zou, Kai and He, Yinan and Zhang, Fan and Gu, Lulu and Zhang, Yuanhan and He, Jingwen and Zheng, Wei-Shi and others},
  journal = {arXiv preprint arXiv:2503.21755},
  year = {2025},
}

@article{bib:open_sora,
  title = {{Open-Sora}: Democratizing Efficient Video Production for All},
  author = {Zheng, Zangwei and Peng, Xiangyu and Yang, Tianji and Shen, Chenhui and Li, Shenggui and Liu, Hongxin and Zhou, Yukun and Li, Tianyi and You, Yang},
  journal = {arXiv preprint arXiv:2412.20404},
  year = {2024},
}

@inproceedings{bib:flovd,
  title = {{FloVD}: Optical Flow Meets Video Diffusion Model for Enhanced Camera-Controlled Video Synthesis},
  author = {Jin, Wonjoon and Dai, Qi and Luo, Chong and Baek, Seung-Hwan and Cho, Sunghyun},
  _booktitle={Proceedings of the IEEE/CVF Conference on Computer Vision and Pattern Recognition (CVPR)},
  booktitle=CVPR,
  year = {2025},
}

@inproceedings{bib:motionctrl,
  title = {{MotionCtrl}: A Unified and Flexible Motion Controller for Video Generation},
  author = {Wang, Zhouxia and Yuan, Ziyang and Wang, Xintao and Li, Yaowei and Chen, Tianshui and Xia, Menghan and Luo, Ping and Shan, Ying},
  _booktitle={Proceedings of the ACM SIGGRAPH Conference (SIGGRAPH)},
  booktitle=SIGGRAPH,
  year = {2024},
}

@article{bib:rays_as_pixels,
  title = {Rays as Pixels: Learning A Joint Distribution of Videos and Camera Trajectories},
  author = {Jang, Wonbong and Liu, Shikun and Sanyal, Soubhik and Perez, Juan Camilo and Ng, Kam Woh and Agrawal, Sanskar and Perez-Rua, Juan-Manuel and Douratsos, Yiannis and Xiang, Tao},
  journal = {arXiv preprint arXiv:2604.09429},
  year = {2026},
}

@inproceedings{bib:motion_attribution,
  title = {Motion Attribution for Video Generation},
  author = {Wu, Xindi and Paschalidou, Despoina and Gao, Jun and Torralba, Antonio and Leal-Taix{\'e}, Laura and Russakovsky, Olga and Fidler, Sanja and Lorraine, Jonathan},
  _booktitle={Proceedings of the International Conference on Machine Learning (ICML)},
  booktitle=ICML,
  year = {2026},
}

@inproceedings{bib:bridgedata_v2,
  title = {{BridgeData V2}: A Dataset for Robot Learning at Scale},
  author = {Walke, Homer and Black, Kevin and Lee, Abraham and Kim, Moo Jin and Du, Max and Zheng, Chongyi and Zhao, Tony and Hansen-Estruch, Philippe and Vuong, Quan and He, Andre and Myers, Vivek and Fang, Kuan and Finn, Chelsea and Levine, Sergey},
  _booktitle={Proceedings of the Conference on Robot Learning (CoRL)},
  booktitle=CoRL,
  year = {2023},
}

@article{bib:astra_mobile_robots,
  title = {Astra: Toward General-Purpose Mobile Robots via Hierarchical Multimodal Learning},
  author = {Chen, Sheng and He, Peiyu and Hu, Jiaxin and Liu, Ziyang and Wang, Yansheng and Xu, Tao and Zhang, Chi and others},
  journal = {arXiv preprint arXiv:2506.06205},
  year = {2025},
}

@article{bib:motus,
  title = {Motus: A Unified Latent Action World Model},
  author = {Bi, Hongzhe and Tan, Hengkai and Xie, Shenghao and Wang, Zeyuan and Huang, Shuhe and Liu, Haitian and Zhao, Ruowen and Feng, Yao and Xiang, Chendong and Rong, Yinze and Zhao, Hongyan and Liu, Hanyu and Su, Zhizhong and Ma, Lei and Su, Hang and Zhu, Jun},
  journal = {arXiv preprint arXiv:2512.13030},
  year = {2025},
}

@article{bib:large_video_planner,
  title = {Large Video Planner Enables Generalizable Robot Control},
  author = {Chen, Boyuan and Zhang, Tianyuan and Geng, Haoran and Song, Kiwhan and Zhang, Caiyi and Li, Peihao and Freeman, William T. and Malik, Jitendra and Abbeel, Pieter and Tedrake, Russ and Sitzmann, Vincent and Du, Yilun},
  journal = {arXiv preprint arXiv:2512.15840},
  year = {2025},
}

@inproceedings{bib:cosmos_policy,
  title = {Cosmos Policy: Fine-Tuning Video Models for Visuomotor Control and Planning},
  author = {Kim, Moo Jin and Gao, Yihuai and Lin, Tsung-Yi and Lin, Yen-Chen and Ge, Yunhao and Lam, Grace and Liang, Percy and Song, Shuran and Liu, Ming-Yu and Finn, Chelsea and Gu, Jinwei},
  _booktitle={Proceedings of the International Conference on Learning Representations (ICLR)},
  booktitle=ICLR,
  year = {2026},
}

@inproceedings{bib:anypoint_trajectory,
  title = {Any-point Trajectory Modeling for Policy Learning},
  author = {Wen, Chuan and Lin, Xingyu and So, John and Chen, Kai and Dou, Qi and Gao, Yang and Abbeel, Pieter},
  _booktitle={Proceedings of Robotics: Science and Systems (RSS)},
  booktitle=RSS,
  year = {2024},
}

@inproceedings{bib:irasim,
  title = {{IRASim}: A Fine-Grained World Model for Robot Manipulation},
  author = {Zhu, Fangqi and Wu, Hongtao and Guo, Song and Liu, Yuxiao and Cheang, Chilam and Kong, Tao},
  _booktitle={Proceedings of the IEEE/CVF International Conference on Computer Vision (ICCV)},
  booktitle=ICCV,
  _pages={9834--9844},
  year = {2025},
}

@inproceedings{bib:iso_dream,
  title = {{Iso-Dream}: Isolating and Leveraging Noncontrollable Visual Dynamics in World Models},
  author = {Pan, Minting and Zhu, Xiangming and Wang, Yunbo and Yang, Xiaokang},
  _booktitle = {Proceedings of Advances in Neural Information Processing Systems (NeurIPS)},
  booktitle = NIPS,
  year = {2022},
}

@inproceedings{bib:droid,
  title = {{DROID}: A Large-Scale In-the-Wild Robot Manipulation Dataset},
  author = {Khazatsky, Alexander and Pertsch, Karl and Nair, Suraj and others},
  _booktitle = {Proceedings of Robotics: Science and Systems (RSS)},
  booktitle = RSS,
  year = {2024},
}

@inproceedings{bib:worldagen,
  title = {{WorldAgen}: Unified State-Action Prediction with Test-Time World Model Training},
  author = {Wan, Chi and Wang, Kangrui and Si, Yuan and Zhang, Pingyue and Li, Manling},
  _booktitle = {Proceedings of the AAAI Conference on Artificial Intelligence (AAAI)},
  booktitle = AAAI,
  pages = {18584--18592},
  year = {2026},
}

@inproceedings{bib:mask2iv,
  title = {Mask2IV: Interaction-Centric Video Generation via Mask Trajectories},
  author = {Li, Gen and Zhao, Bo and Yang, Jianfei and Sevilla-Lara, Laura},
  _booktitle = {Proceedings of the AAAI Conference on Artificial Intelligence (AAAI)},
  booktitle = AAAI,
  year = {2026},
}

@inproceedings{bib:sam3,
  title = {{SAM 3}: Segment Anything with Concepts},
  author = {Carion, Nicolas and Gustafson, Laura and Hu, Yuan-Ting and Debnath, Shoubhik and Hu, Ronghang and Suris Coll-Vinent, Didac and Ryali, Chaitanya and Alwala, Kalyan Vasudev and Khedr, Haitham and Huang, Andrew and Lei, Jie and Ma, Tengyu and Guo, Baishan and Kalla, Arpit and Marks, Markus and Greer, Joseph and Wang, Meng and Sun, Peize and R{\"a}dle, Roman and Afouras, Triantafyllos and Mavroudi, Effrosyni and Xu, Katherine and Wu, Tsung-Han and Zhou, Yu and Momeni, Liliane and Hazra, Rishi and Ding, Shuangrui and Vaze, Sagar and Porcher, Francois and Li, Feng and Li, Siyuan and Kamath, Aishwarya and Cheng, Ho Kei and Doll{\'a}r, Piotr and Ravi, Nikhila and Saenko, Kate and Zhang, Pengchuan and Feichtenhofer, Christoph},
  _booktitle = {Proceedings of the International Conference on Learning Representations (ICLR)},
  booktitle = ICLR,
  year = {2026},
}

@misc{bib:gemini25,
  title = {Gemini 2.5: Pushing the Frontier with Advanced Reasoning, Multimodality, Long Context, and Next Generation Agentic Capabilities},
  author = {{Gemini Team, Google}},
  year = {2025},
}
